\documentclass[10pt,twocolumn]{article}
\usepackage[utf8]{inputenc}
\usepackage{simpleConference}
\usepackage{times}
\usepackage{latexsym}
\usepackage{graphicx}
\usepackage{subfigure}
\usepackage{multirow}
\usepackage{amssymb}
\usepackage{amsmath}
\usepackage[linesnumbered,ruled,vlined]{algorithm2e}
\usepackage{url,hyperref}
\usepackage{balance}
\usepackage[table]{xcolor}
\usepackage{cite}
\usepackage{epstopdf}
\usepackage{makecell}
\usepackage[T1]{fontenc}

\SetKwInput{KwInput}{Input}      
\SetKwInput{KwOutput}{Output}    

\title{Improving Global Adversarial Robustness Generalization\\With Adversarially Trained GAN}

\author{Desheng Wang\textsuperscript{1}\and Weidong Jin\textsuperscript{1}\and Yunpu Wu\textsuperscript{1}\and Aamir Khan\textsuperscript{1}}

\begin{document}

\maketitle

\let\thefootnote\relax\footnotetext{\noindent\textsuperscript{1}School of Electrical Engineering, Southwest Jiaotong University, Chengdu 611756, P.~R.~China. Correspondence to: Desheng Wang \\\textless \textit{wds@my.swjtu.edu.cn}\textgreater}

\begin{abstract}
Convolutional neural networks (CNNs) have achieved beyond human-level accuracy in the image classification task and are widely deployed in real-world environments. However, CNNs show vulnerability to adversarial perturbations that are well-designed noises aiming to mislead the classification models. In order to defend against the adversarial perturbations, adversarially trained GAN (ATGAN) is proposed to improve the adversarial robustness generalization of the state-of-the-art CNNs trained by adversarial training. ATGAN incorporates adversarial training into standard GAN training procedure to remove obfuscated gradients which can lead to a false sense in defending against the adversarial perturbations and are commonly observed in existing GANs-based adversarial defense methods. Moreover, ATGAN adopts the image-to-image generator as data augmentation to increase the sample complexity needed for adversarial robustness generalization in adversarial training. Experimental results in MNIST SVHN and CIFAR-10 datasets show that the proposed method doesn't rely on obfuscated gradients and achieves better global adversarial robustness generalization performance than the adversarially trained state-of-the-art CNNs.
\end{abstract}

\section{Introduction}
\label{sec:1}
Deep learning \cite{1} has achieved great success on a range of computer vision applications such as image classification \cite{2,3,4,5,6,7}, object detection \cite{8,9,10}, and semantic segmentation \cite{11,12,13}. However, recent researches \cite{14,15} demonstrate that Deep Neural Networks are vulnerable to adversarial perturbations. The adversarial perturbations are well-designed noises that can for example mislead a well-trained CNN to output incorrect class labels with high confidence. The perturbations are imperceptible to humans and can transfer across different model parameters even architectures. The examples combining the clean examples with the adversarial perturbations are so-called adversarial examples. The adversarial examples post a great threat to the Deep Learning-based security-crucial applications such as self-driving cars \cite{16}, person detection systems \cite{17} or medical diagnosis systems. 

Adversarial examples attract great attention in the deep learning community. Lots of works emerge on the generation of more powerful adversarial examples \cite{14,15,18,19,20,21,22,23,24}. Besides, there are also large amounts of works on how to train the deep learning models to be robust against the adversarial examples \cite{24,15,25,26,27,28,29,30,31,32,33,34,35,36,37,38,39}. A model's ability to behave correctly under adversarial examples is called adversarial robustness. Adversarial robustness is denoted by the accuracy of the classification model on the adversarial examples. One of the lines of research on the adversarial defense is generative model-based adversarial defense methods \cite{36,37,38,39}. However, these methods are shown to lead to a false sense of security in defending against adversarial examples due to obfuscated gradients \cite{40} or evaluated under weak threat models \cite{41}. Therefore it is significant to remove the obfuscated gradients of the generative model-based adversarial defense methods and make them robust under strong adversarial attacks. Adversarial training (AT) \cite{24,15} is by so far the only effective adversarial defense method which doesn't rely on the obfuscated gradients and can truly improve the adversarial robustness of the target model. Nevertheless, the CNN classifiers trained by AT still suffer from the poor robustness generalization problem especially for complicated datasets, and this robustness generalization gap is further enlarged when the perturbations larger than the value used during training \cite{24}.

Recently, Schmidt \textit{et al.} \cite{42} studied adversarially robust learning from the generalization perspective. According to their theoretical analysis, the overfitting of the classifiers trained by AT is due to the lack of training data. The sample complexity of adversarial robustness generalization is significantly larger than that of standard generalization.

A common solution to improve the standard generalization when deep learning works with limited data is to use data augmentation \cite{43}. The commonly used data augmentation techniques are geometric transformations such as flipping, cropping, rotation, and translation which are demonstrated can improve the standard generalization accuracy of CNN classifiers on CIFAR-10, CIFAR-100, and ImageNet datasets \cite{4,7}. The above-mentioned data augmentation methods can only simply manipulate the original data in pixel space. The Generative Adversarial Networks (GANs) \cite{54}, due to the ability to model the data distribution and generate new training data, have already been adopted as high-level data augmentation technique and resulted in better performance classification models \cite{44,45}. For the robustness generalization we study in this paper, Liu \textit{et al.} \cite{46} demonstrated that the robustness of classifiers trained by AT can be improved if we have a deeper understanding of the image data distribution. And adding a generator to the AT procedure is shown to have the benefit of improving the robustness of the discriminator. Therefore, the GANs can be used as the data augmentation technique to improve the robustness generalization of the CNN classifiers trained by AT.

In order to defend against the adversarial examples, ATGAN is proposed in this paper which can simultaneously get rid of the obfuscated gradients and improve the adversarial robustness generalization performance of the CNN classifiers trained by AT. In the proposed ATGAN framework, the min-max AT procedure \cite{24} is incorporated into the standard GANs training procedure. Experimental results show that ATGAN achieves better adversarial robustness generalization performance than the state-of-the-art CNN classifiers trained by AT. Our contributions are listed as follows:

\begin{description}

  \item[$\bullet$] A novel adversarial defense method called ATGAN is proposed with the combination of the cGANs and the AC-GANs. Moreover, the perceptual loss is adopted to increase the quality of the output of the generator.
  \item[$\bullet$] AT is incorporated into the standard GANs training procedure. This combination not only makes the proposed method gets rid of obfuscated gradients but also increase the sample complexity needed by adversarial robustness generalization.
  \item[$\bullet$] Extensive experiments are conducted to evaluate the proposed method on MNIST SVHN and CIFAR-10 datasets. The experimental results demonstrate that the proposed method achieves better adversarial robustness generalization performance against the $L_{\infty}$ norm bounded adversarial examples than the state-of-the-art CNN classifiers trained by AT.

\end{description}

The rest of the paper is arranged as follows: In Section \ref{sec:2}, we review some related works about the adversarial attack and defense methods, the high sample complexity for adversarial robustness generalization, and the generative adversarial networks. In Section \ref{sec:3}, we introduce our proposed ATGAN method. In Section \ref{sec:4}, we list the experimental results to demonstrate the feasibility of ATGAN. In Section \ref{sec:5}, we do some discussions about our experimental results. Finally, we give some conclusions in Section \ref{sec:6}.

\section{Related Work}
\label{sec:2}
\subsection{Adversarial Attacks and Defenses}
\label{sec:2.1}
Since Szegedy \textit{et al.} \cite{14} first discovered the vulnerability of neural networks to the adversarial examples, researches have proposed various of adversarial attacks methods to generate adversarial examples for evaluating the worst-case robustness of the neural network models. These works generate small perturbations under $L_{p}$ norm constrains, such as $L_{0}$ norm, $L_{2}$ norm or $L_{\infty}$ norm. For the $L_{0}$ norm constrain, Papernot \textit{et al.} \cite{19} proposed JSMA method, Su \textit{et al.} \cite{20} proposed one pixel attack method. For the $L_{2}$ norm constrain, Carlini and Wagner \cite{21} proposed C\&W attack method, Moosavi-Dezfooli at al. \cite{22} proposed DeepFool method. For the $L_{\infty}$ norm, Szegedy \textit{et al.} \cite{14} proposed L-BFGS method, Goodfellow at al. \cite{15} proposed FGSM method, Madry \textit{et al.} \cite{24} proposed PGD method.

Many adversarial defense methods have been proposed to make the neural network models robust to the $L_{p}$ norm constrained adversarial examples. These works include modifying the input such as data compression \cite{26} and feature squeezing \cite{35}, modifying the network such as stochastic activation pruning \cite{47} and Deep Contractive Network \cite{28}, adding auxiliary network such as Defense-gan \cite{37}, Pixeldefend \cite{36}, and Magnet \cite{38}. Nevertheless, these methods are demonstrated to fail to defend against the adversarial examples. AT \cite{24} trains the neural network models on the adversarial examples in a min-max manner, which is by so far the only defense method that can improve the adversarial robustness of the neural networks.

\subsection{High Sample Complexity for Adversarial Robustness Generalization}
\label{sec:2.2}
Although AT can improve the adversarial robustness of CNNs, the CNNs trained by AT still suffer from the problem of poor adversarial robustness generalization performance. Schmidt \textit{et al.} \cite{42} point out that adversarial robustness generalization requires much more labeled data than standard generalization. Based on this theoretical work, a follow-up work done by Zhai \textit{et al.} \cite{48} demonstrates that the labeled sample complexity for adversarial robustness generalization can be largely reduced if unlabeled data is used, and an adversarially robust model which generalizes well can be learned if we have plenty of unlabeled data. Another independent work done by Uesato \textit{et al.} \cite{49} also shows that an Unsupervised Adversarial Training (UAT) algorithm which uses unlabeled data together with very limited labeled data can significantly improve the adversarial robustness generalization performance achieved by purely supervised approaches. Carmon \textit{et al.} \cite{50} show that unlabeled data can fill the sample complexity gap between adversarial robustness generalization and standard generalization through a self-training algorithm. Najafi \textit{et al.} \cite{51} extent the Distributionally Robust Learning (DRL) method to handle Semi-Supervised Learning (SSL). Their proposed method demonstrates a comparable performance to that of the state-of-the-art method. Besides, two interesting works \cite{52,53} show that self-supervised learning can improve the adversarial robustness of the CNNs.

\subsection{Generative Adversarial Networks}
\label{sec:2.3}
Generative Adversarial Networks (GANs) are an approach to generative modeling using deep learning methods such as CNNs. The idea of GANs was first introduced by Goodfellow \textit{et al.} \cite{54} in 2014. The architecture of GANs consists of a generator that aims to generate photo-realistic images from randomly selected noise vectors, and a discriminator that aims to discriminate between real images and generated images. The generator and the discriminator are trained simultaneously in a zero-sum game. Although the impressive results achieved by GANs, the training process of GANs is unstable. In \cite{55}, the authors give an architecture guideline for designing both the generator and the discriminator, which is demonstrated can result in stable GANs models. Based on these works, plenty of methods that extend GANs on the architecture or the loss function are proposed and achieve state-of-the-art results on various applications. Among these methods, the most related ones to our work are conditional GANs (cGANs) for image-to-image translation \cite{56} and Auxiliary Classifier GANs (AC-GANs) \cite{57}. The cGANs learn to map from conditional information $c$ and random noise vector $z$ to commonly an image $y$, $G:\{c, z\}\rightarrow y$, which is different from the traditional GANs that learn a mapping from random noise vector $z$ to output image $y$, $G:z\rightarrow y$. In \cite{56}, the authors use images as the conditional information and leverage an image-to-image network to generate synthetic images from images of other domain. The AC-GANs modify the discriminator to contain an auxiliary classifier that outputs the class label for the training data. The experimental results show that the GANs tasked with classification objective result in higher quality samples.

\section{Method}
\label{sec:3}
ATGAN is proposed to defend against the $L_{\infty}$ norm bounded adversarial examples from the perspective of mapping the adversarial examples back onto the clean examples manifold. The image-to-image network is widely adopted as the generator in cGANs to translate the images from one domain to another. Thus ATGAN uses the image-to-image network generator to map the adversarial examples back onto the clean examples manifold. However, the GANs trained by the standard training procedure have obfuscated gradients in the generator, leading to a false sense on defending against the adversarial examples. AT is by so far the only adversarial defense method that doesn't rely on the obfuscated gradients and can truly improve the adversarial robustness generalization of the CNNs. In order to remove the obfuscated gradients in GANs, ATGAN incorporates AT into the standard GANs training procedure. AT is a supervised training algorithm while the GANs is an unsupervised method, thus AT cannot be used to train the GANs directly. Inspired by AC-GANs, ATGAN adds an auxiliary decoder to the discriminator that is tasked with reconstructing class labels. During training, the discriminator distinguishes the output of the generator from the real images and classify them simultaneously. Therefore, ATGAN can be treated as a classifier, and AT can be applied to ATGAN directly without modification. Moreover, the perceptual loss is adopted to improve the perceptual quality and the sample complexity of the output of the generator, which can further improve the adversarial robustness generalization.

\subsection{GAN Adversarial Training}
\label{sec:3.1}
The main idea of ATGAN is incorporating AT into the standard GANs training procedure in order to remove the obfuscated gradients in GANs. Formally, a standard classification task can be defined as finding the set of model parameters $\theta\in\mathbb{R}^p$ that minimize the empirical risk $\mathbb{E}_{(x,y)\sim D}[L(x,y,\theta)]$ where $x\in\mathbb{R}^d$ and $y\in[k]$ are pairs of examples and corresponding class labels sampled from the empirical data distribution $D$ and $L(x,y,\theta)$ is a loss function such as the cross-entropy loss for a neural network. Although yielding classifiers with high standard generalization accuracy, empirical risk minimization (ERM) doesn't ensure the robustness to adversarial examples. AT extends ERM through robust optimization. In AT, a threat model is specified which defines the attacks the classifiers should be robust to. During training, instead of directly feeding into the loss $L$ the samples from the distribution $D$, AT allows the threat model to perturb the samples first. This can be formalized as the min-max optimization problem:
\begin{equation}
\label{equation1}
\min_{\theta}\mathbb{E}_{(x,y)\sim D}[\max_{\delta\in S}L(x+\delta,y,\theta)]
\end{equation}
where $S\subseteq\mathbb{R}^d$ is the set of perturbations the threat model can find, such as the union of the $L_{\infty}$-balls around the clean examples $x$. In this min-max optimization, the inner maximization problem finds an adversarial example that maximizes the loss, which is solved by projected gradient descent (PGD) \cite{24}:
\begin{equation}
\label{equation2}
x^{t+1}=\prod_{x+S}(x^t+\alpha\mathrm{sgn}(\nabla_{x}L(x,y,\theta)))
\end{equation}
The outer minimization problem finds model parameters so that the loss is minimized on the adversarial examples found by the inner maximization, which can be solved by back-propagation for neural networks. AT doesn't rely on obfuscated gradients when defending against the adversarial examples.
GANs are also adopted as adversarial defense method based on the idea of mapping the adversarial examples back onto the manifold of the clean examples. However, the GANs trained by the standard training procedure have obfuscated gradients that give a false sense on the adversarial robustness. In order to remove the obfuscated gradients in GANs, making GANs truly robust against adversarial examples, ATGAN incorporates AT into the standard GANs training procedure. For ATGAN, the inner maximization problem in the min-max optimization \ref{equation1} is modified as:
\begin{equation}
\label{equation3}
\max_{\delta\in S}L(G(x+\delta),y,\theta)
\end{equation}
where $G$ is the image-to-image network generator of ATGAN. Before fed into the classifier, the samples are first mapped by the generator. During the inner maximization, the generator combined with the classifier, or the discriminator here, is treated as a whole classifier which the threat model aims to generate the adversarial examples against. This operation not only guarantees that ATGAN can learn the mapping from the distribution of the adversarial examples to the manifold of the clean examples but also guarantees that the obfuscated gradients in the generator can be removed.

\subsection{Adversarial Robustness Generalization}
\label{sec:3.2}
In order to improve the adversarial robustness generalization performance of the CNNs trained by AT, ATGAN uses data augmentation to increase the sample complexity of the adversarial examples. The sample complexity needed by adversarial robustness generalization is higher than that needed by standard generalization. Thus the training data used to train a model with high standard generalization accuracy often cannot yield a model of similar high adversarial robustness generalization accuracy. The image-to-image network generator of ATGAN first encodes the adversarial examples into the low-dimensionality latent space then decodes the latent vectors into the clean example manifold. This operation can reduce the magnitude of the perturbations in the adversarial examples thus can reduce the power of the adversarial examples. ATGAN also adopts the perceptual loss which uses a pre-trained CNN to extract the high-level feature representations, then calculates the difference of the output of the generator and the clean examples based on these high-level feature representations:
\begin{equation}
\label{equation4}
L_{p}^{\phi,j}(\hat{x},x)=\frac{1}{C_jH_jW_j}\Vert\phi_j(\hat{x})-\phi_j(x)\Vert^{2}_{2}
\end{equation}
where $\phi_j$ denotes the activations of the feature map of shape $C_j\times H_j\times W_j$ which are extracted from the $j$th layer of the pre-trained CNN, $\Vert\cdot\Vert_{2}$ denotes the Euclidean distance between feature representations. The perceptual loss can drive the output of the generator to be perceptually similar to the clean examples but not match the output of the generator and the clean examples exactly. Therefore the generator plays s role of data augmentation that can increase the sample complexity of the adversarial examples. As a result, ATGAN improves the adversarial robustness generalization performance of the CNNs trained by AT.

\begin{figure*}[!htb]
  \centering
  \subfigure[]{\includegraphics[width=5.7cm]{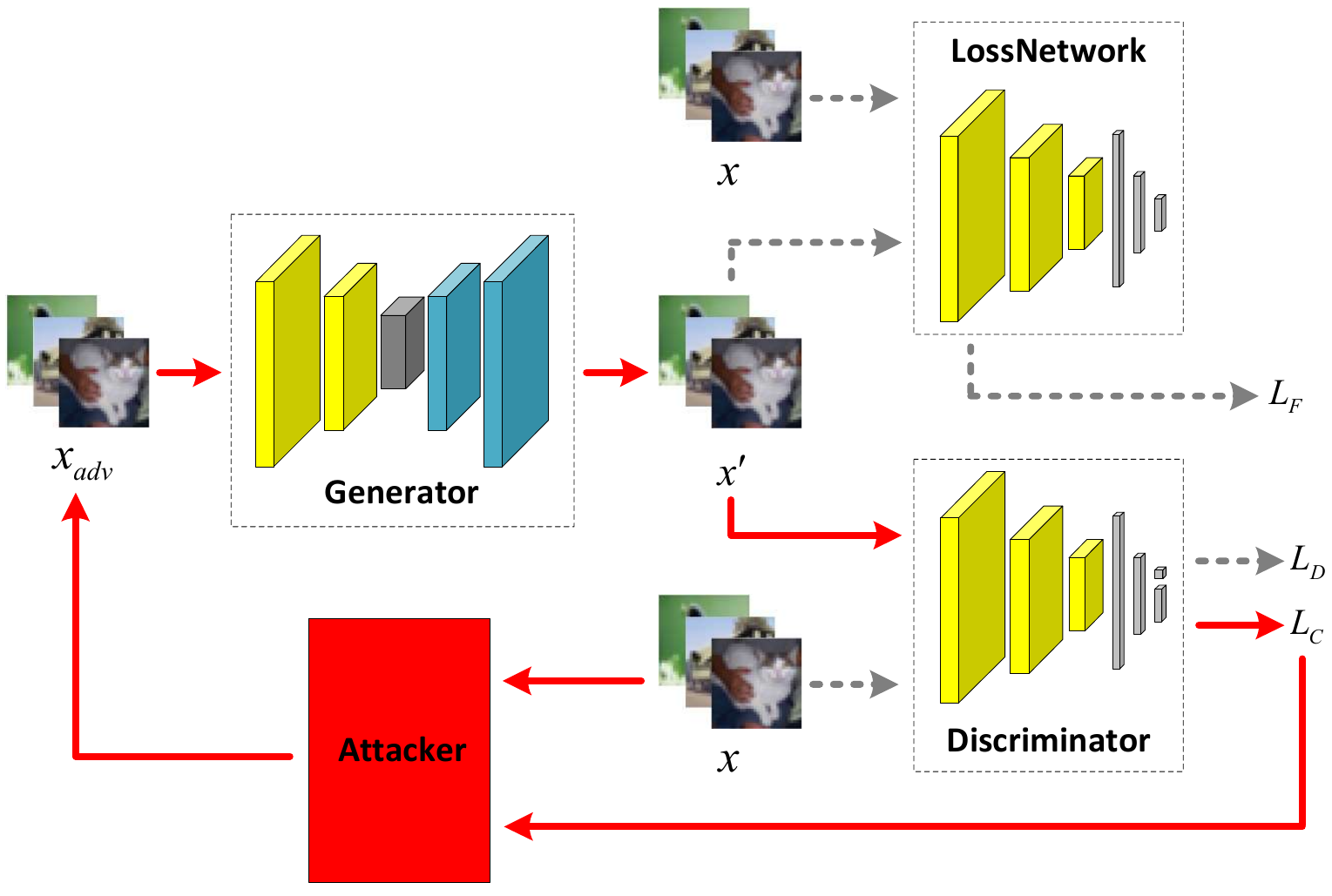}}
  \subfigure[]{\includegraphics[width=5.7cm]{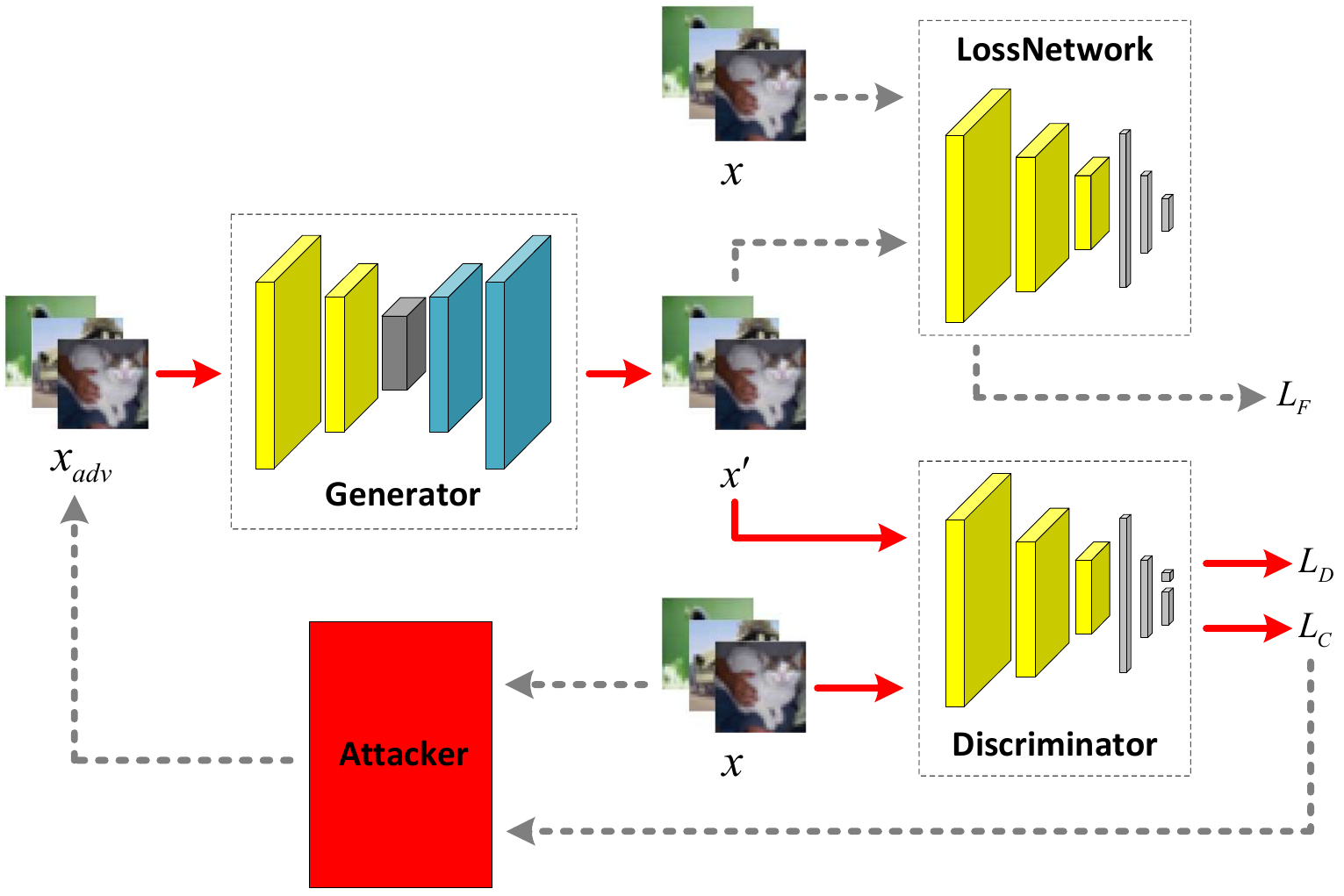}}
  \subfigure[]{\includegraphics[width=5.7cm]{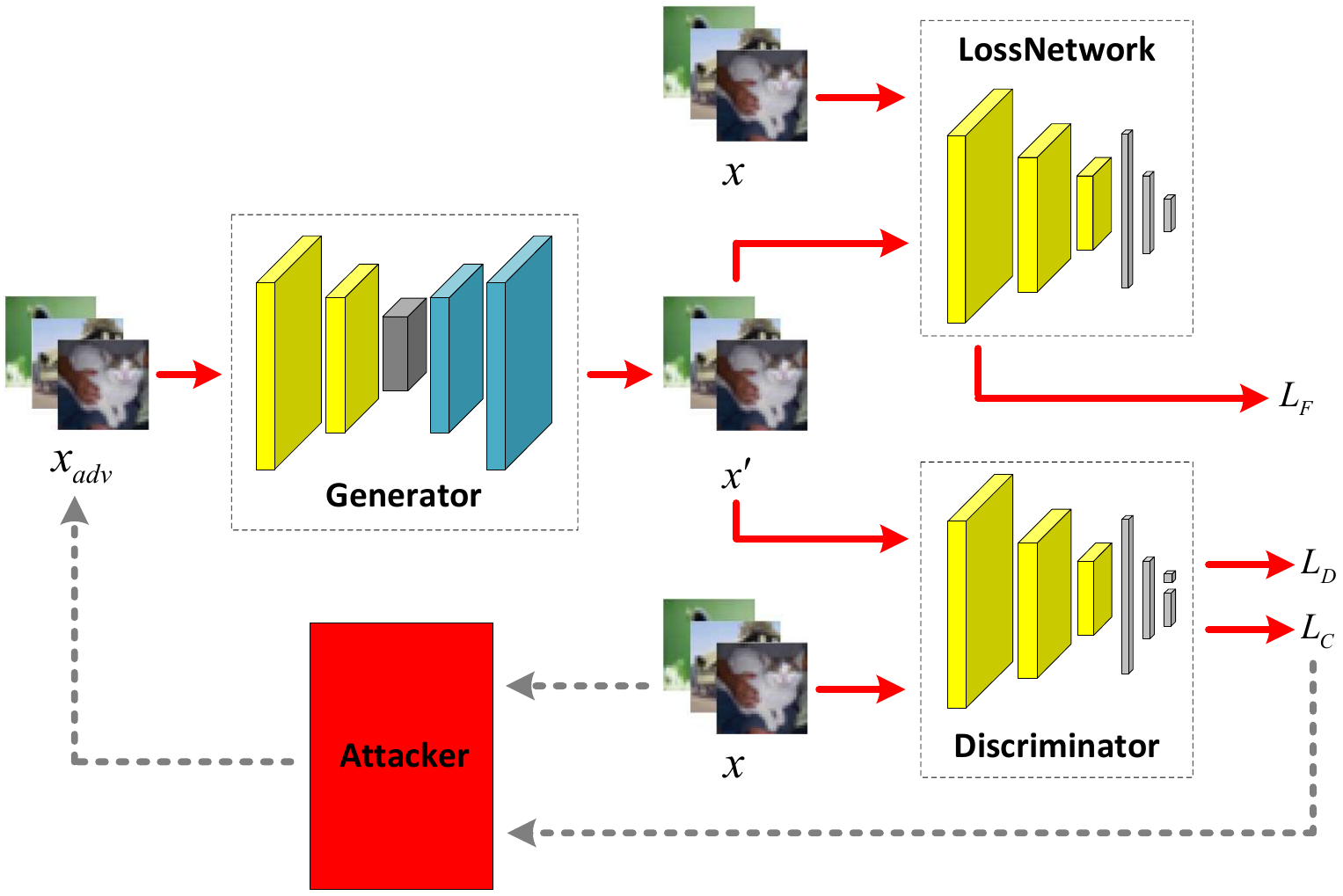}}
  \caption{\small The architecture and training procedure of ATGAN. (a) The inner maximization loop of the AT. (b) The training of the discriminator (the outer minimization loop). (c) The training of the generator (the outer minimization loop). The red solid arrows indicate the path of the data and the calculation of the loss value in the corresponding procedure. The gray dotted arrows indicate that there is no data flow along with them in this procedure}
  \label{Fig1}
\end{figure*}

\begin{table*}[!htb]
  \centering
  \caption{\small The architectures of the generators of ATGAN for MNIST SVHN and CIFAT-10}
  \label{table1}
  \begin{tabular}{c|c|c}
    \hline
    MNIST                     & SVHN                                 & CIFAR-10                                      \\ \hline
    \makecell[c]{Conv(8,3,1)\\BN\\Relu}\Bigg\}$\times$1       & \makecell[c]{Conv(64,4,2)\\BN\\LeakyRelu(0.2)}\Bigg\}$\times$1       & \makecell[c]{Conv(128,4,2)\\BN\\LeakyRelu(0.2)}\Bigg\}$\times$3      \\
    \makecell[c]{Conv(16,3,2)\\BN\\Relu}\Bigg\}$\times$1      & \makecell[c]{Conv(128,4,2)\\BN\\LeakyRelu(0.2)}\Bigg\}$\times$1      & \makecell[c]{TransConv(128,4,2)\\BN\\LeakyRelu(0.2)}\Bigg\}$\times$3 \\
    \makecell[c]{Conv(32,3,2)\\BN\\Relu}\Bigg\}$\times$1      & \makecell[c]{TransConv(128,4,2)\\BN\\LeakyRelu(0.2)}\Bigg\}$\times$1 & Conv(3,3,1) \\
    ResBlock(32)$\times$4                                     & \makecell[c]{TransConv(64,4,2)\\BN\\LeakyRelu(0.2)}\Bigg\}$\times$1  & tanh        \\
    \makecell[c]{TransConv(16,3,2)\\BN\\Relu}\Bigg\}$\times$1 & Conv(3,3,1)                                                          &             \\
    \makecell[c]{TransConv(8,3,2)\\BN\\Relu}\Bigg\}$\times$1  & tanh                                                                 &             \\
    Conv(3,3,1)                                               &                                                                      &             \\
    tanh                                                      &                                                                      &             \\
    \hline
  \end{tabular}
\end{table*}

\subsection{ATGAN Architecture}
\label{sec:3.3}
The schematic of the architecture of ATGAN is exhibited in Fig. \ref{Fig1}. ATGAN consists of an adversarial attacker denoted by A, a generator denoted by G, a discriminator denoted by D, and a loss network denoted by L. The adversarial attacker is the $L_{\infty}$ norm PGD described in section \ref{sec:3.1} which treats the whole network consisted of the generator and the discriminator as a classifier and attacks this classifier to generate adversarial examples in one training epoch according to formula \ref{equation2}, indicated by Fig. \ref{Fig1} (a). The generated adversarial examples and their clean examples counterparts are then used to train the generator and the discriminator during the same training epoch. The generator is a fully convolutional image-to-image network that maps the adversarial examples onto the clean examples manifold. The strided convolutions and deconvolutions are adopted to downsample and upsample the adversarial examples instead of pooling layers. Tanh nonlinearity is adopted for the output layer to bound the values of the pixels in the range $[0,1]$. The exact architectures and hyper-parameters of the generators are different for different datasets, which are shown in Table \ref{table1}. ATGAN adopts an AC-GANs-like discriminator which has an auxiliary classifier tasked with reconstructing the class labels of the inputs. The discrimination output of the discriminator produces a scalar value between 0 and 1 that indicates how likely the input image is a clean image. The classification output of the discriminator produces a $k$-dimensional vector that indicates the discrete probability distribution of $k$ classes. The discriminator is used to discriminate the adversarial examples and the clean examples and predict their class labels simultaneously. The discrimination task can drive the generator to output images that are similar to the clean examples and the classification task can further improve the perceptual quality of the output images. The CNN used in \cite{21}, the DenseNet40 \cite{59} and the ResNet18 \cite{60} are adopted as discriminators respectively for MNIST SVHN and CIFAR-10 datasets. ATGAN adopts a pre-trained CNN as the loss network and the activations of the first convolutional layers of the loss network are used to calculate the perceptual loss for the feature representations extracted from early layers of the loss network tends to encourage the generator to produce images that are visually realistic. The architectures of the loss network are the same with the discriminators for MNIST SVHN and CIFAR-10 datasets, except that there is no discrimination output.

\subsection{Loss Functions}
\label{sec:3.4}
ATGAN incorporates AT into standard GANs training procedure to remove obfuscated gradients and improve the adversarial robustness generalization performance of the CNN classifiers trained by AT. Thus the loss functions for both AT and GANs are extended by ATGAN.

\subsubsection{Adversarial Loss}
\label{sec:3.4.1}
To incorporate the AT into standard GANs training procedure, the generator G and the discriminator D are first treated as an entire classifier (the discrimination output of the discriminator is ignored in this step), and the PGD is used as the adversarial attacker A to generate a set of adversarial examples $P$ on this classifier by solving problem \ref{equation5}:
\begin{equation}
\begin{split}
\label{equation5}
P=\{x_{adv}|x_{adv}=x+\delta, \delta=\mathop{\arg\max}_{\Vert\delta\Vert_{\infty}\leq\epsilon}\\L_{adv}(\theta, x+\delta, y), (x, y)\sim P_{data}\}
\end{split}
\end{equation}
where $L_{adv}$ is the adversarial loss:
\begin{equation}
\label{equation6}
L_{adv}(\theta, x+\delta, y)=c(D_c(G(x+\delta;\theta_G);\theta_D),y)
\end{equation}
where $c(\cdot)$ denotes the cross-entropy, $\theta=\{\theta_G,\theta_D\}$, $\theta_G$ denotes the weights and biases of the generator, $\theta_D$ denotes the weights and the biases of the discriminator, and $D_c(\cdot)$ denotes the classification output of the discriminator. The adversarial attacker finds the perturbations that maximize the adversarial loss. During this step, $\theta_D$ and $\theta_G$ are fixed. This inner maximization loop of the AT is shown in Fig. \ref{Fig1} (a).

\subsubsection{Discriminator Loss}
\label{sec:3.4.2}
After generating the adversarial examples against the intermediate snapshot of the ATGAN model, the parameters of ATGAN are updated by one step on the adversarial examples and their clean examples counterpart. This includes the updating of the parameters of the discriminator and the generator. 
The discriminator is trained by solving problem \ref{equation7}:
\begin{equation}
\label{equation7}
\theta_D=\mathop{\arg\min}_{\theta_D}(L_c(x,x_{adv},y;\theta)-L_d(x,x_{adv};\theta))
\end{equation}
where $L_c$ and $L_d$ are the classification loss and the discriminator loss:
\begin{equation}
\begin{split}
\label{equation8}
L_c(x,x_{adv},y;\theta)=E_{(x,y)\sim P_{data}}c(D_C(x;\theta_D),y)+\\E_{y\sim P_{data},x_{adv}\in P}c(D_C(G(x_{adv};\theta_G);\theta_D),y)
\end{split}
\end{equation}
\begin{equation}
\begin{split}
\label{equation9}
L_d(x,x_{adv};\theta)=E_{x\sim P_{data}}\log D_D(x;\theta_D)+\\E_{x_{adv}\in P}\log(1-D_D(G(x_{adv};\theta_G);\theta_D))
\end{split}
\end{equation}
where $D_D(\cdot)$ is the scalar discrimination output of the discriminator. The first term and the second term in the classification loss \ref{equation8} indicate the losses on the clean examples and the adversarial examples. For convenience, we denote them as $L_{a}$ and $L_{b}$.

\subsubsection{Generator Loss}
\label{sec:3.4.3}
The generator is trained by solving problem \ref{equation10}:
\begin{equation}
\begin{split}
\label{equation10}
\theta_G=\mathop{\arg\min}_{\theta_G}(L_c(x,x_{adv},y;\theta)-\\L_g(x_{adv};\theta)+L_f(x,x_{adv};\theta_G))
\end{split}
\end{equation}
where $L_g$ and $L_f$ are the generator loss and the perceptual loss:
\begin{equation}
\label{equation11}
L_g(x_{adv};\theta)=E_{x_{adv}\in P}\log D_D(G(x_{adv};\theta_G);\theta_D)
\end{equation}
\begin{equation}
\begin{split}
\label{equation12}
L_f(x,x_{adv};\theta_G)=E_{x\sim P_{data},x_{adv}\in P}\frac{1}{C_jH_jW_j}\\\Vert\phi_j(G(x_{adv};\theta_G))-\phi_j(x)\Vert^{2}_{2}
\end{split}
\end{equation}
where $\phi_j(\cdot)$ denotes the activations of the feature map of the $j$th layer of the loss network of shape $C_j\times H_j\times W_j$, $\Vert\cdot\Vert_{2}$ denotes the Euclidean distance between feature representations.

\subsubsection{Full Objective}
\label{sec:3.4.4}
The full objective to train ATGAN is presented as below:
\begin{equation}
\label{equation13}
L_{adv}
\end{equation}
\begin{equation}
\label{equation14}
L_{D}=\alpha(L_{a}+\beta L_{b})-L_{d}
\end{equation}
\begin{equation}
\label{equation15}
L_{G}=\alpha(L_{a}+\beta L_{b})-L_{g}+\gamma L_{f}
\end{equation}
The $\alpha$, $\beta$ and, $\gamma$ in equation \ref{equation14} and \ref{equation15} are weight parameters to control the importance of different loss items. These three steps are repeated until the model converges. The training algorithm of ATGAN is described in Algorithm \ref{algorithm1}.
\begin{algorithm}
\label{algorithm1}
\SetAlgoLined
 \KwInput{$(x,y)\sim P_{data}$, clean examples; $\theta_D^0$, initial discriminator parameters; $\theta_G^0$, initial generator parameters; $\alpha_1$, $\alpha_2$, $\beta$, and $\gamma$, loss weights}
 \KwOutput{$\theta_D$, $\theta_G$}
 \While{$\theta_D$ and $\theta_G$ have not converged}
 {
  Sample a batch from the training examples: $\{(x^i,y^i)\}_{i=1}^m\sim P_{data}$\;
  \For{$i=1,2,...,m$}
  {
   $\delta^i=\mathop{\arg\max}_{\Vert\delta\Vert_{\infty}\leq\epsilon}L_{adv}(\theta, x^i+\delta, y^i)$\;
   $x_{adv}^i=x^i+\delta^i$\;
  }
  $\theta_D=\mathop{\arg\min}_{\theta_D}(\alpha_1L_C(x,x_{adv},y;\theta)-L_D(x,x_{adv};\theta))$\;
  $\theta_G=\mathop{\arg\min}_{\theta_G}(\alpha_2L_C(x,x_{adv},y;\theta)-L_G(x_{adv};\theta)+\gamma L_F(x,x_{adv};\theta_G))$\;
 }
\caption{Training of ATGAN}
\end{algorithm}

\subsection{Differences from Similar Works}
\label{sec:3.5}
It is noted that there are several important differences between ATGAN and the previous similar works \cite{46,39}. The Rob-GAN \cite{46} also incorporates AT into the standard GANs training procedure, but it just adversarially trains the discriminator, which is different from ATGAN where the AT is applied to the whole classifier consisted of the generator and the discriminator. The generator of the Rob-GAN takes a noise vector and a class label as input, but the inputs of the generator of ATGAN are adversarial examples. The APE-GAN \cite{39} adversarially trains the GAN, but the procedure of generating the adversarial examples is not incorporated into the GANs training procedure. They generate the adversarial examples on the target classifier they aim to defend then use the generated adversarial examples to train the GAN. The discriminator of the APE-GAN doesn't task with classification objective. And both Rob-GAN and APE-GAN don't use the perceptual loss.

\section{Experiments and Results}
\label{sec:4}
\subsection{Datasets}
\label{sec:4.1}
We evaluate ATGAN on three standard datasets: MNIST SVHN and CIFAR-10. MNIST is a dataset of handwritten digits containing 70000 $28\times28$ gray images in 10 classes. The training set contains 60000 samples and the testing set contains 10000 samples. SVHN is a dataset containing 10 classes of street view house numbers RGB images of size $32\times32$. The training set contains 73257 samples and the testing set contains 26032 samples. CIFAR-10 is a dataset containing 60000 $32\times32$ RGB images in 10 classes. The training set contains 50000 samples and the testing set contains 10000 samples.

\subsection{Threat Model}
\label{sec:4.2}
ATGANs and the baseline models are trained by AT to resist the $L_{\infty}$ norm bounded adversarial examples using PGD. The number of steps of PGD is set to be 40 for MNIST and 7 for SVHN and CIFAR-10. The step size of PGD $\alpha$ is set to be 1 for MNIST and 2/255 for SVHN and CIFAR-10. To compare the global adversarial robustness generalization performance of ATGANs and the baseline models, we set a range of magnitudes of the perturbations during training and testing which are denoted by $\epsilon_t$ and $\epsilon_i$. We choose $\epsilon_t$ from the range [0.1, 0.15, 0.2, 0.25, 0.3] and $\epsilon_i$ from the range [0.1, 0.15, 0.2, 0.25, 0.3, 0.35, 0.4, 0.45, 0.5] for MNIST. We choose $\epsilon_t$ from the range [2/255, 4/255, 6/255, 8/255, 10/255] and $\epsilon_i$ from the range [2/255, 4/255, 6/255, 8/255, 10/255, 12/255, 14/255, 16/255, 18/255, 20/255] for SVHN and CIFAR-10. In addition to the standard AT procedure, we adopt another AT procedure which is of more generality: only use a fraction of adversarial examples in each mini-batch, and the rest of each mini-batch is replaced with clean examples. We denote this method by FracAT.

\subsection{Baseline Models}
\label{sec:4.3}
Three baseline models--a plain CNN \cite{21}, DenseNet40 \cite{59}, and ResNet18 \cite{60}--are trained on MNIST SVHN and CIFAR-10. For comparison, the architectures of the baseline models are set to be the same with the discriminators of ATGANs but discard the discrimination outputs.

\subsection{Training Parameters}
\label{sec:4.4}
On MNIST we use Adam optimizer, learning rate of 0.001, batch size of 128, and epochs of 50 for both the generator and the discriminator. The loss weights are set as $\alpha=1$, $\beta=1$, and $\gamma=10$. On SVHN we use batch size of 64 and epochs of 40 for both the generator and the discriminator. The Adam optimizer parameterized with learning rate of 0.0002, $\beta_1$ of 0.5, and $\beta_2$ of 0.999 is used for the generator, and the Momentum optimizer parameterized with initial learning rate of 0.1 and momentum of 0.9 is used for the discriminator, the learning rate of the discriminator decays by a factor of 10 at epoch 20 and 30. The loss weights are set as $\alpha=1$, $\beta=1$, and $\gamma=10$. On CIFAR-10 we use batch size of 128 and steps of 80000 for both the generator and the discriminator. The Adam optimizer parameterized with learning rate of 0.0002, $\beta_1$ of 0.5, and $\beta_2$ of 0.999 is used for the generator, and the Momentum optimizer parameterized with initial learning rate of 0.1 and momentum of 0.9 is used for the discriminator, the learning rate of the discriminator decays by a factor of 10 at epoch 100 and 150. The loss weights are set as $\alpha=5$, $\beta=4$, and $\gamma=10$.  

\subsection{Evaluation Metrics}
\label{sec:4.5}
The recently proposed robustness curves \cite{61} is used to quantitatively evaluate the global robustness of different models. Accuracy is adopted instead of the error to plot the robustness curves in this paper. According to Risse \textit{et al.} \cite{62}, the commonly adopted scaler perturbation threshold fails to capture important global robustness properties, thus is not sufficient to reliably and meaningfully compare the robustness of different models. The saliency maps \cite{63} is used to qualitatively evaluate the robustness of the features learned by different models.

\begin{figure*}[!htb]
  \centering
  \subfigure{\includegraphics[width=5.6cm]{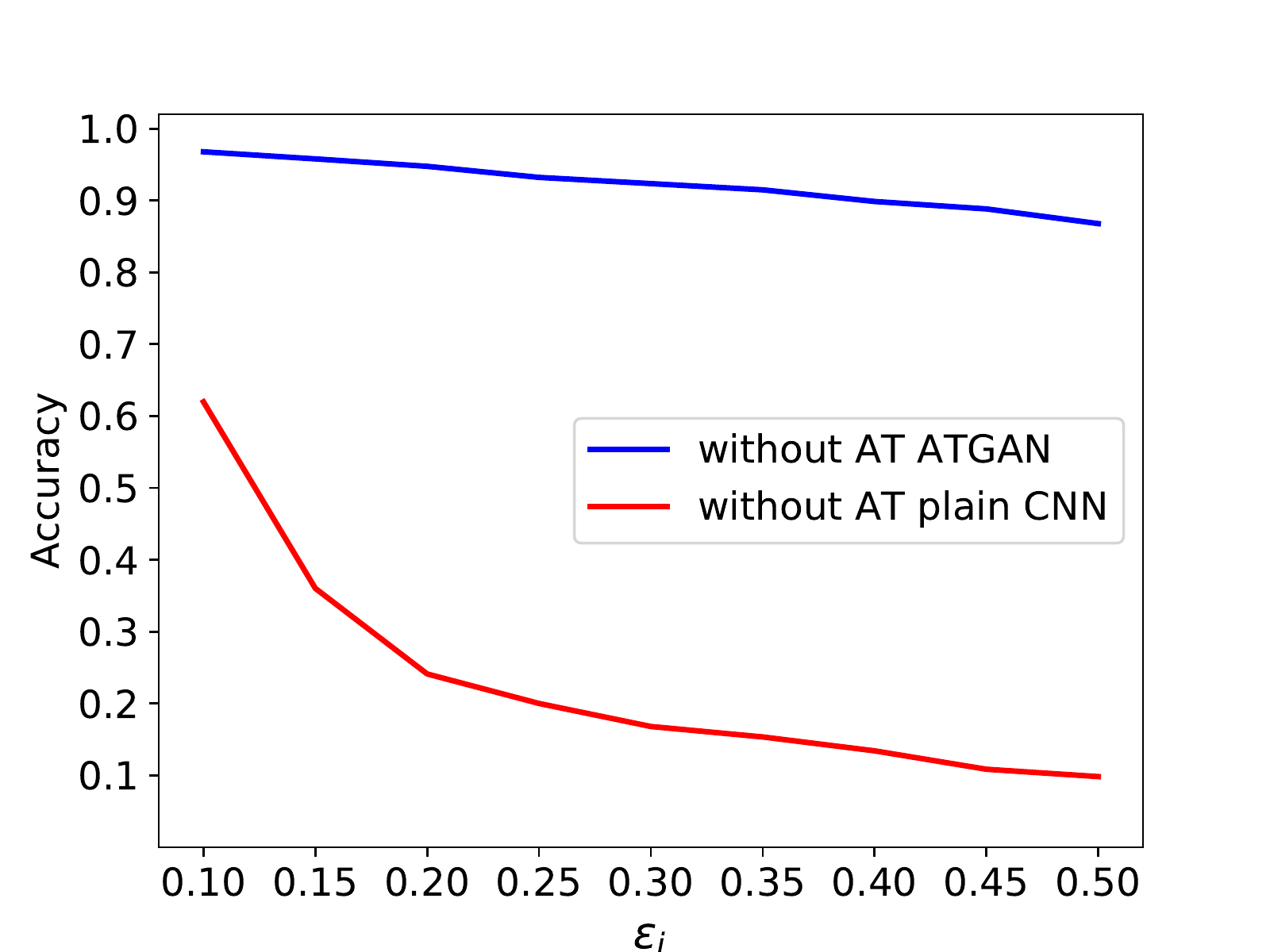}}
  \subfigure{\includegraphics[width=5.6cm]{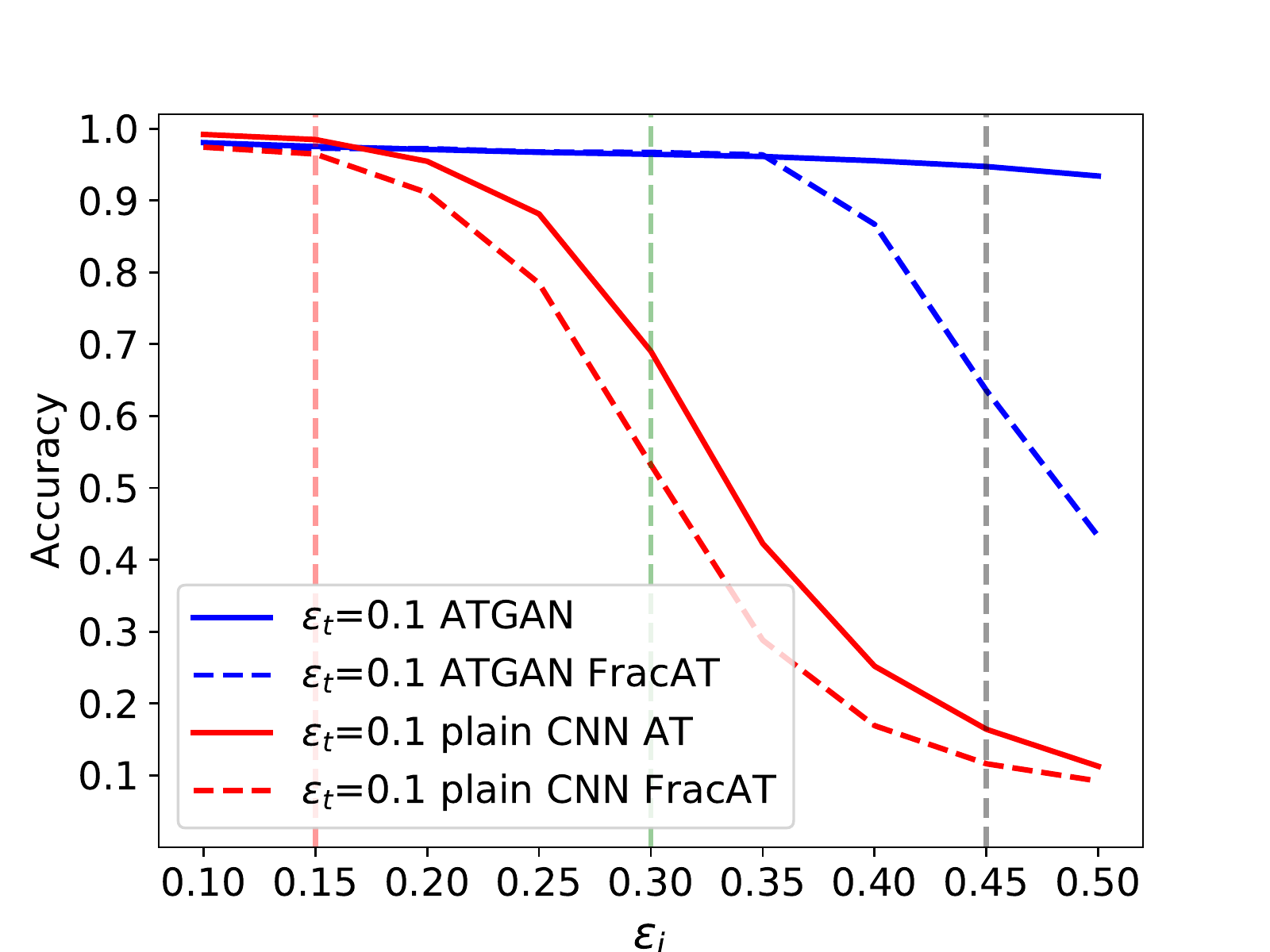}}
  \subfigure{\includegraphics[width=5.6cm]{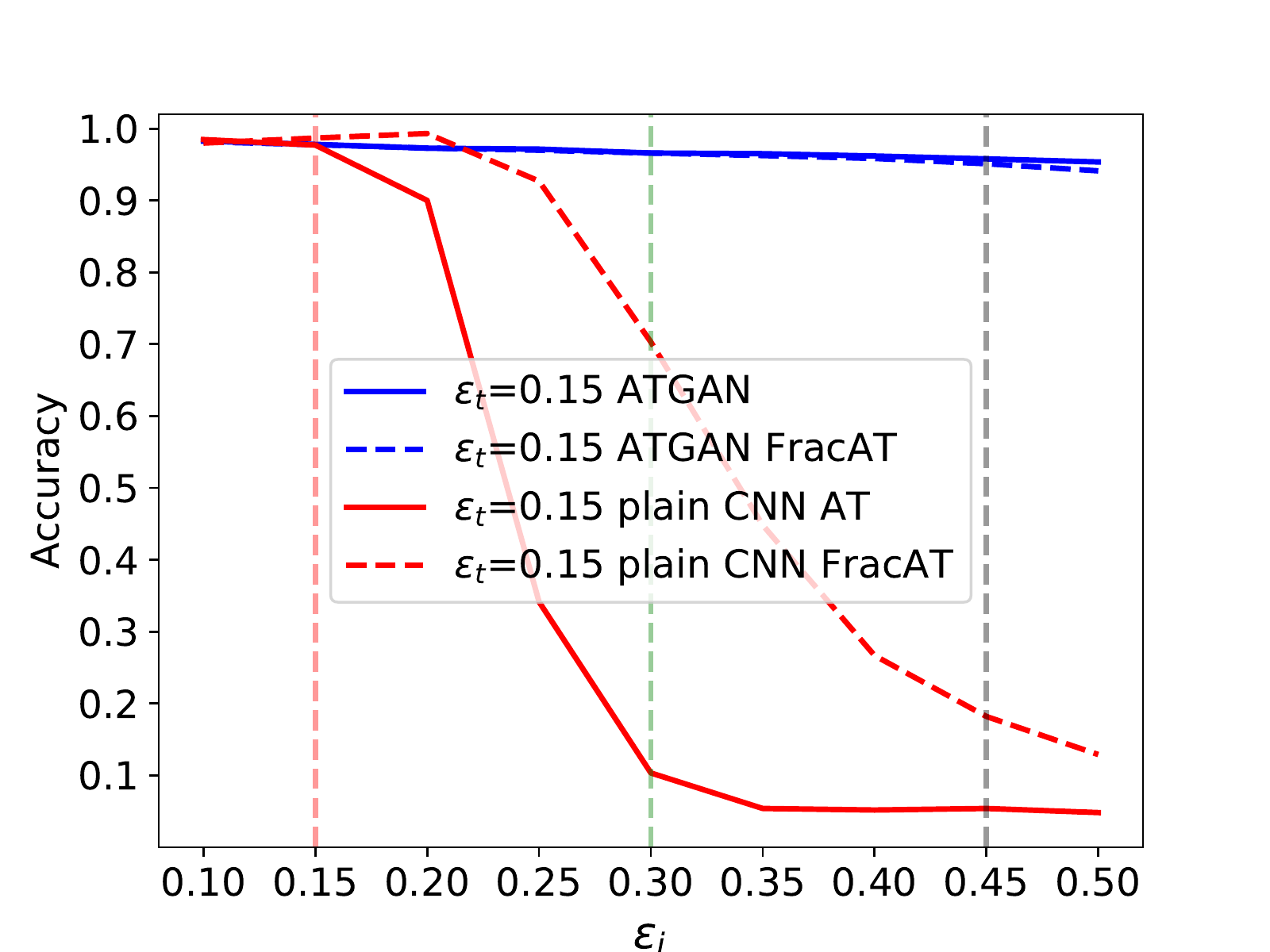}}
  \subfigure{\includegraphics[width=5.6cm]{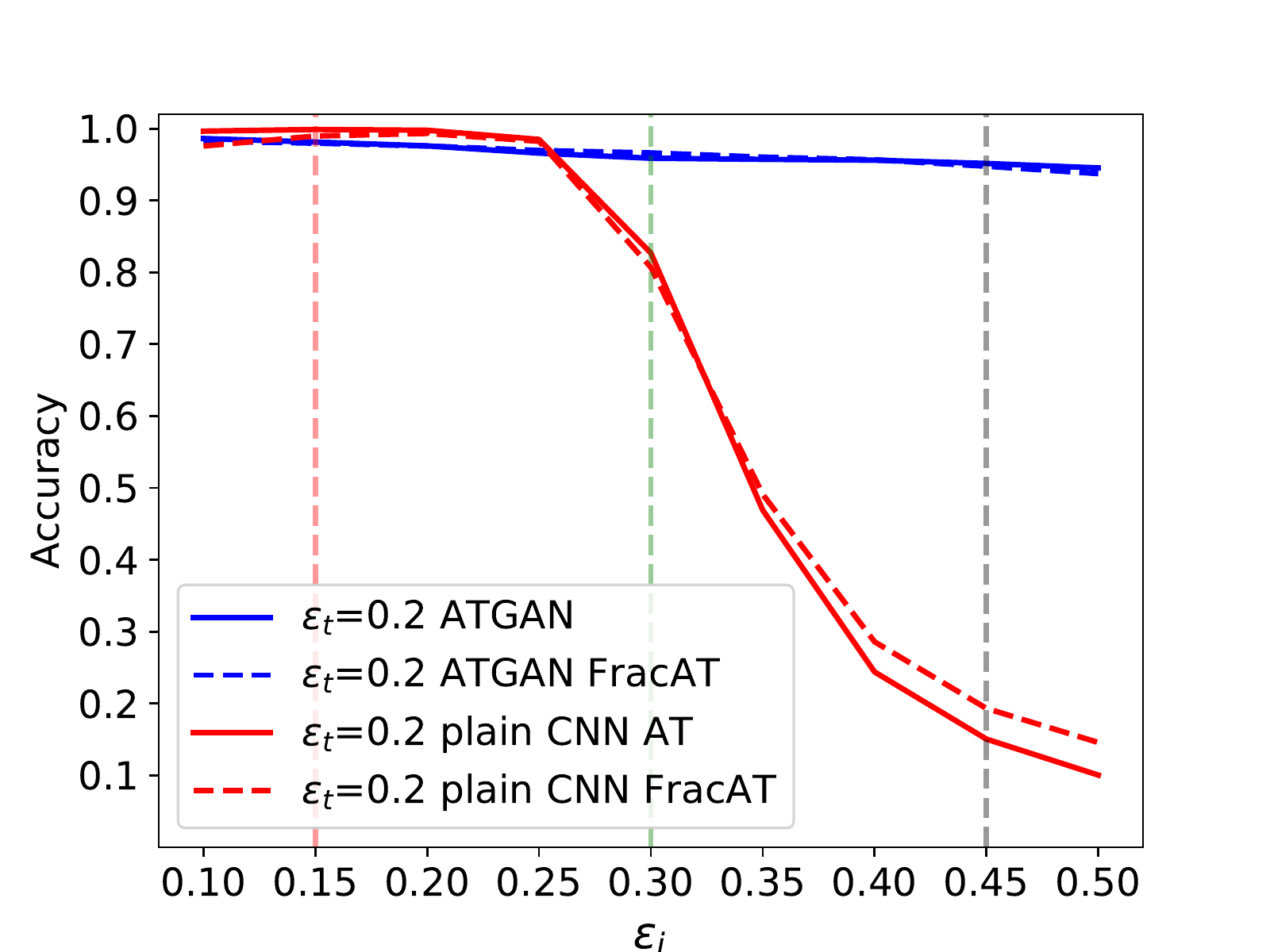}}
  \subfigure{\includegraphics[width=5.6cm]{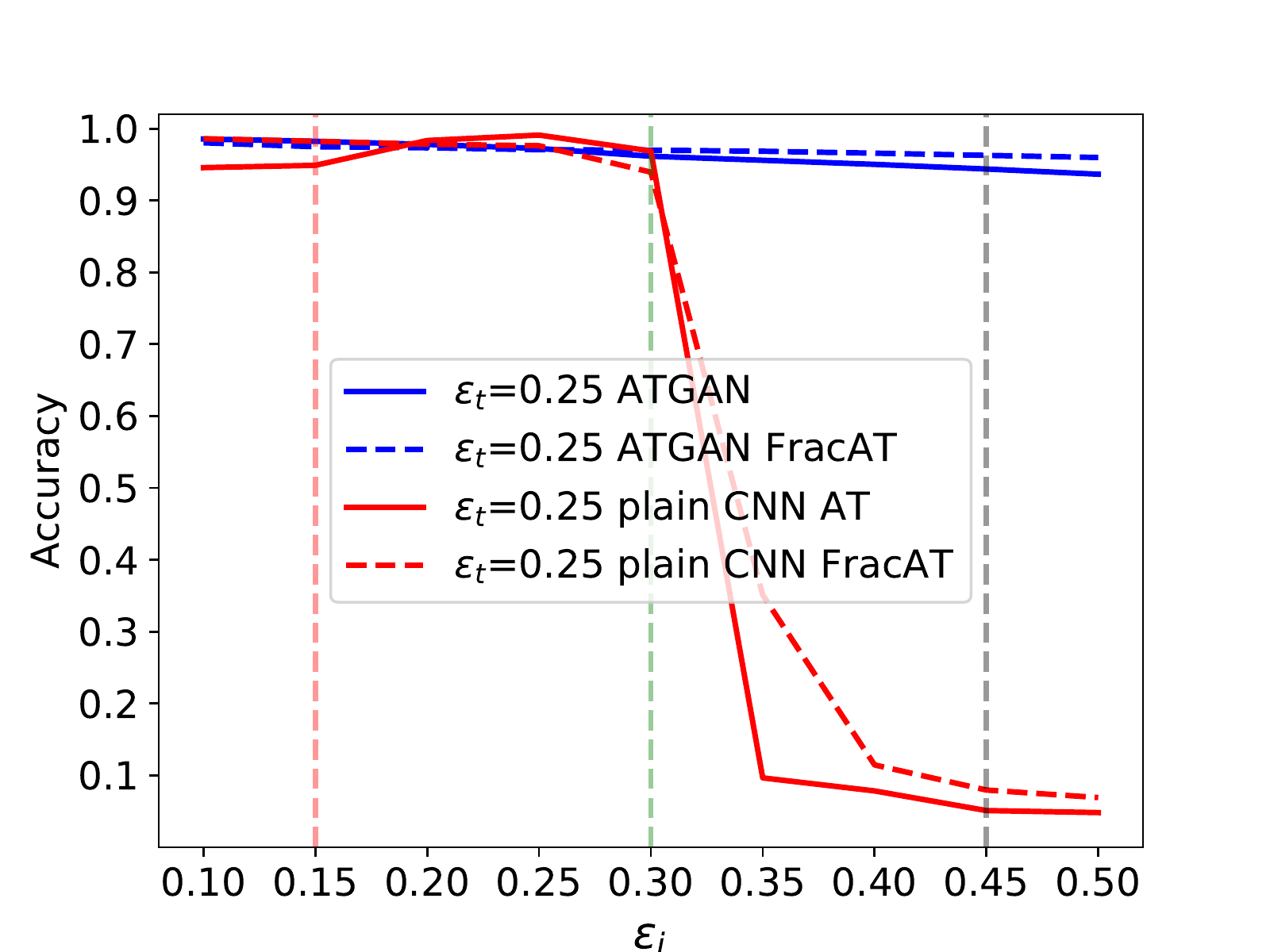}}
  \subfigure{\includegraphics[width=5.6cm]{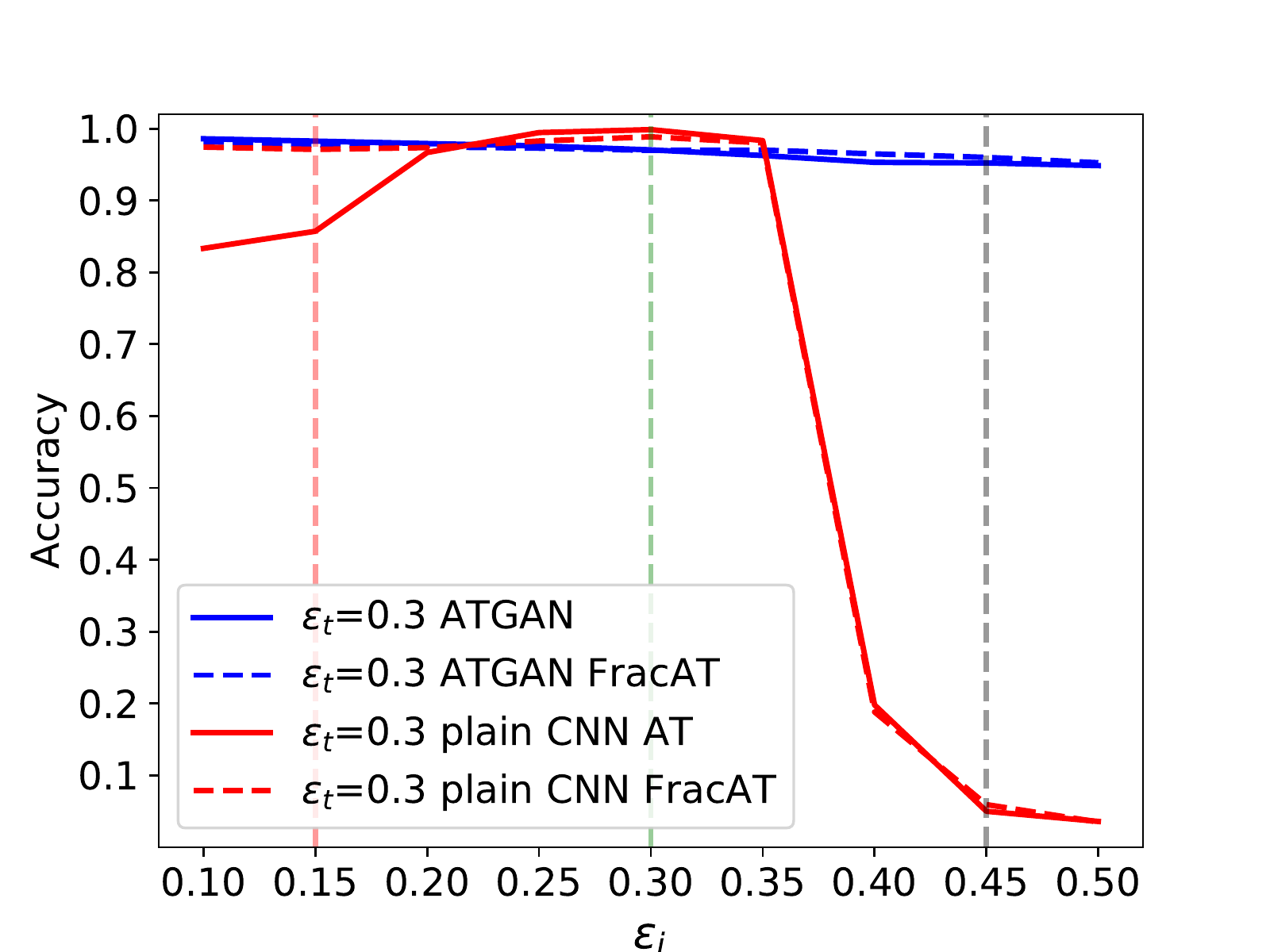}}
  \caption{\small The adversarial robustness generalization performance of ATGAN and plain CNN on MNIST under white-box attack. The solid lines denote ATGAN. The dot-dash lines denote plain CNN}
  \label{Fig2}
\end{figure*}

\begin{figure*}[!htb]
  \centering
  \subfigure{\includegraphics[width=5.6cm]{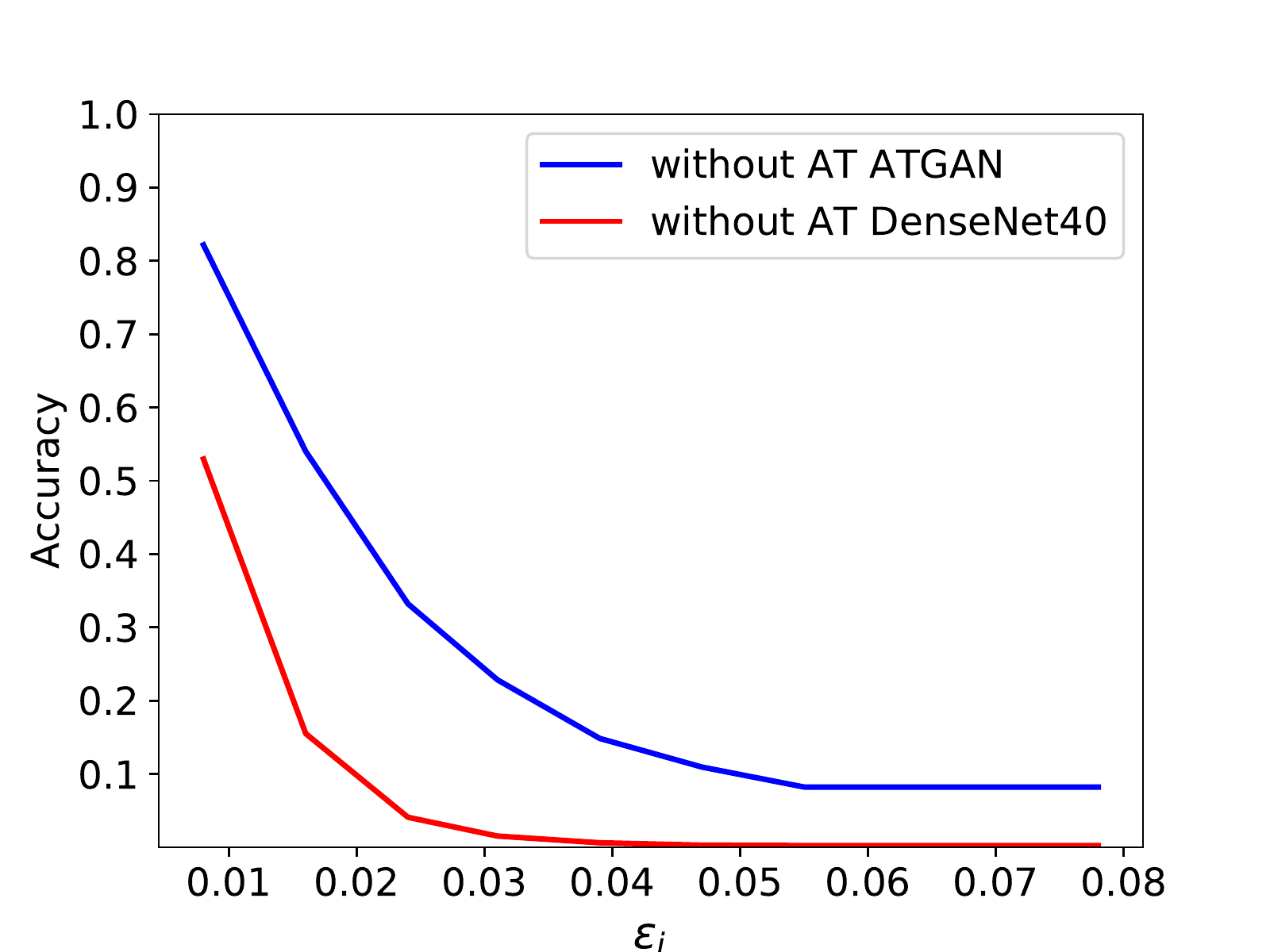}}
  \subfigure{\includegraphics[width=5.6cm]{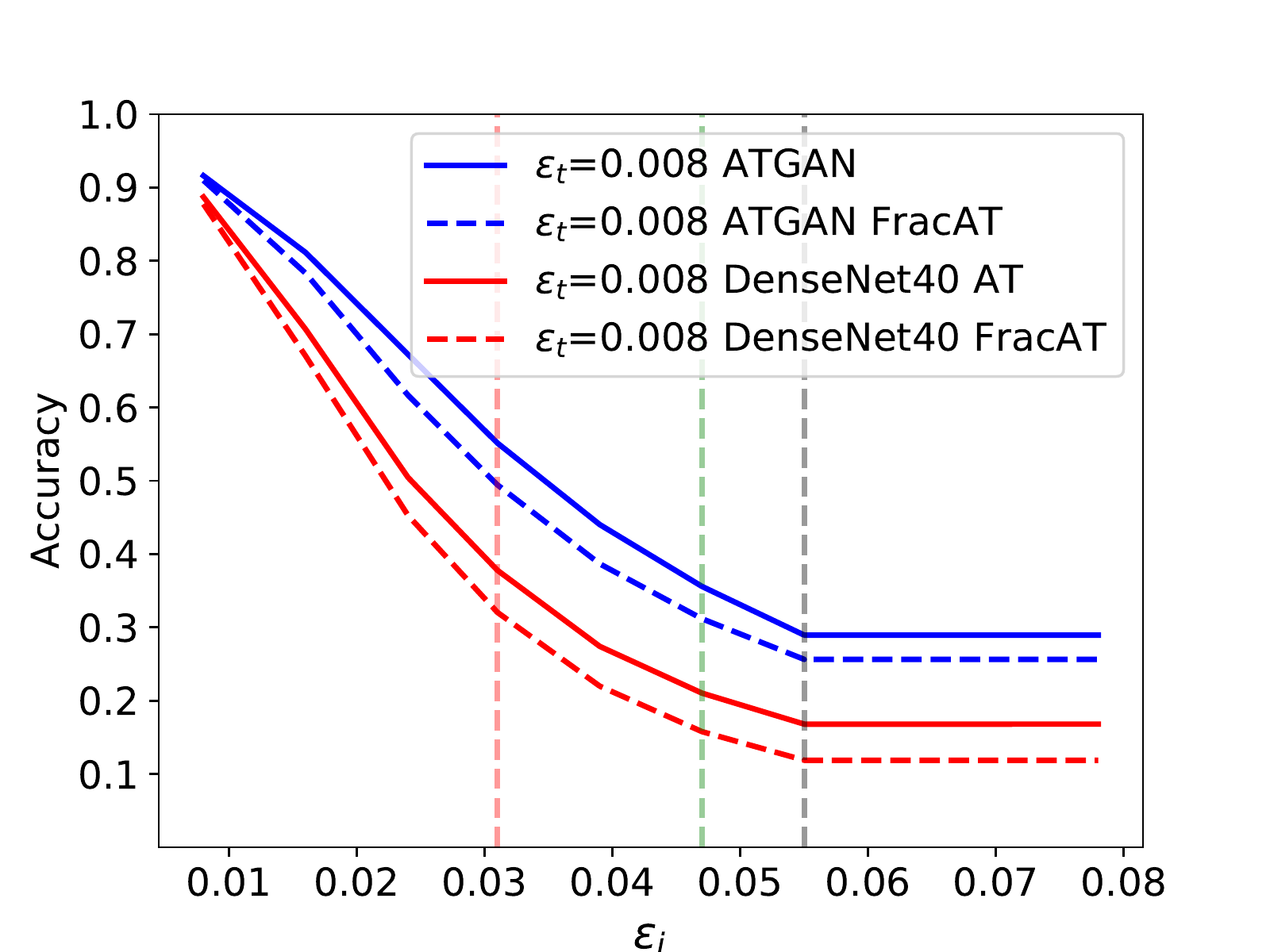}}
  \subfigure{\includegraphics[width=5.6cm]{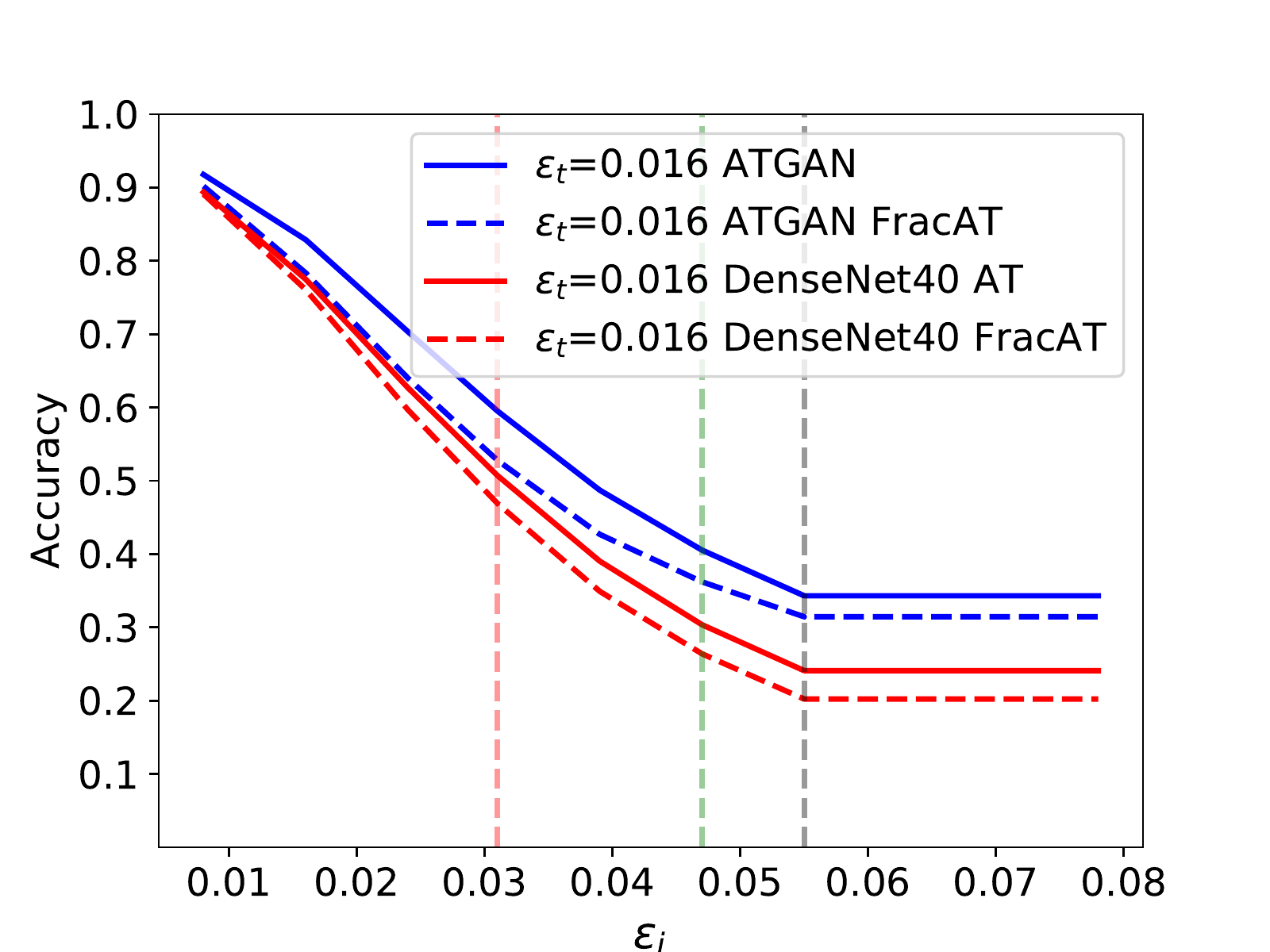}}
  \subfigure{\includegraphics[width=5.6cm]{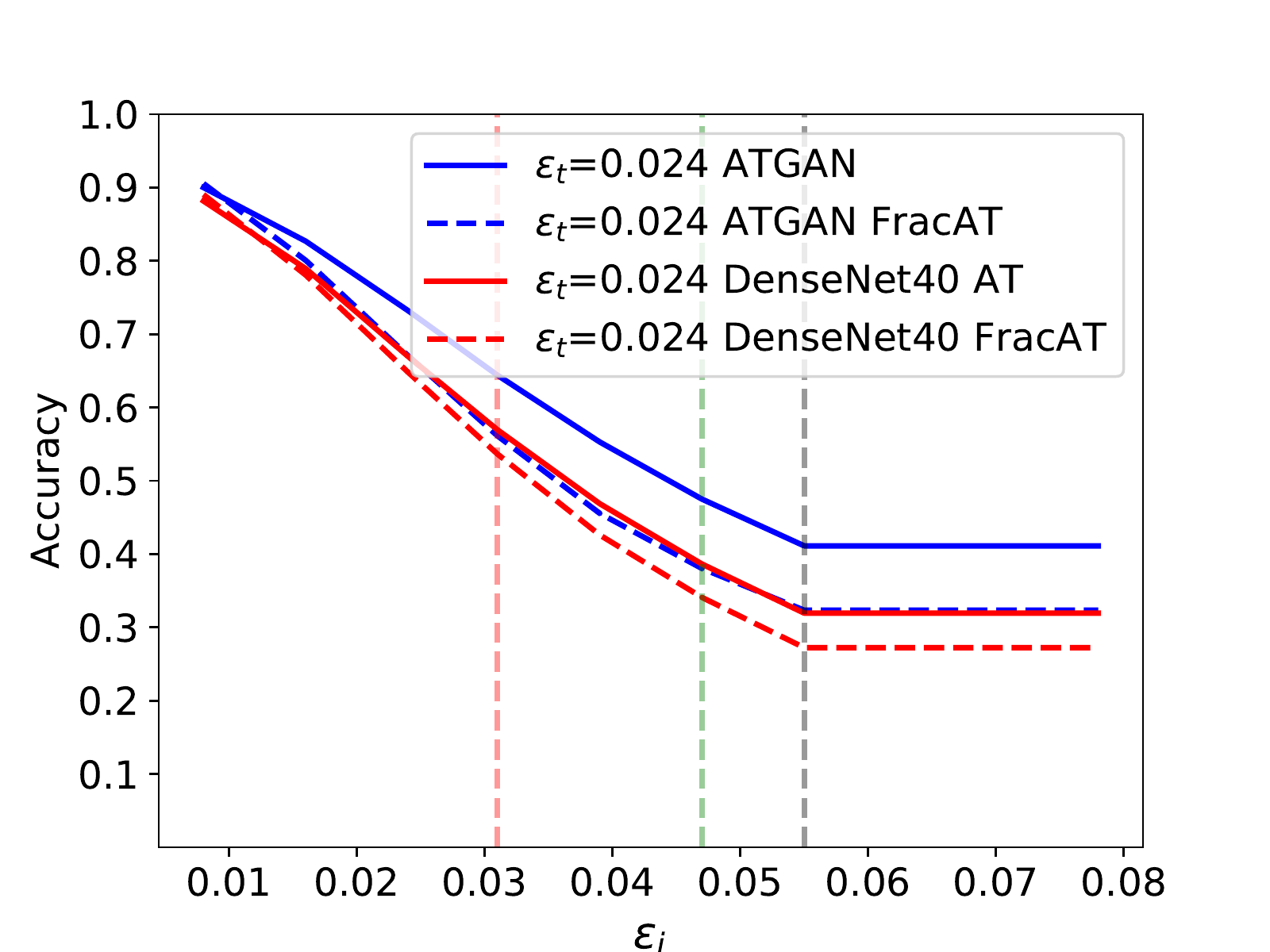}}
  \subfigure{\includegraphics[width=5.6cm]{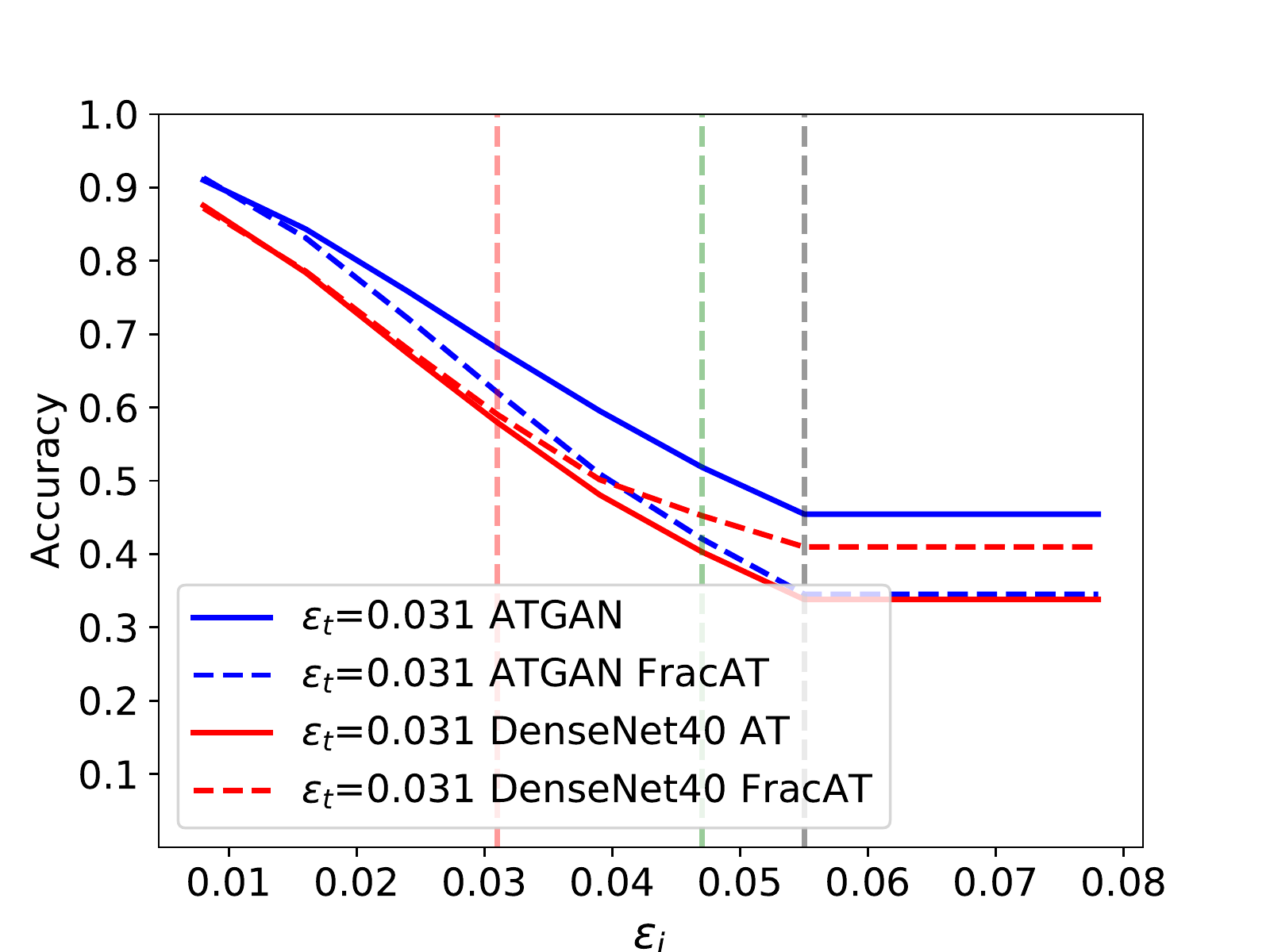}}
  \subfigure{\includegraphics[width=5.6cm]{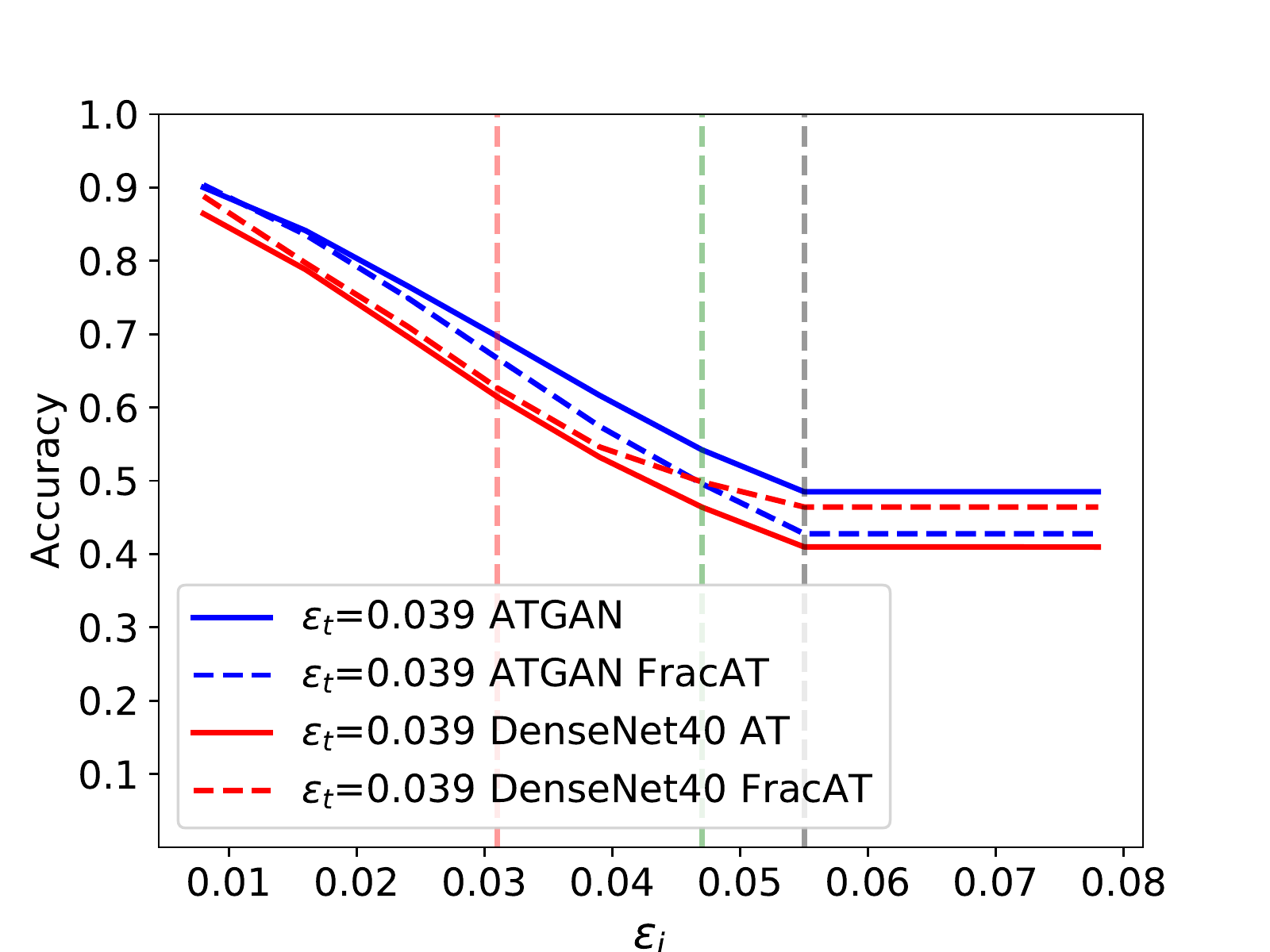}}
  \caption{\small The adversarial robustness generalization performance of ATGAN and DenseNet40 on SVHN under white-box attack. The solid lines denote ATGAN. The dot-dash lines denote DenseNet40}
  \label{Fig3}
\end{figure*}

\begin{figure*}[!htb]
  \centering
  \subfigure{\includegraphics[width=5.6cm]{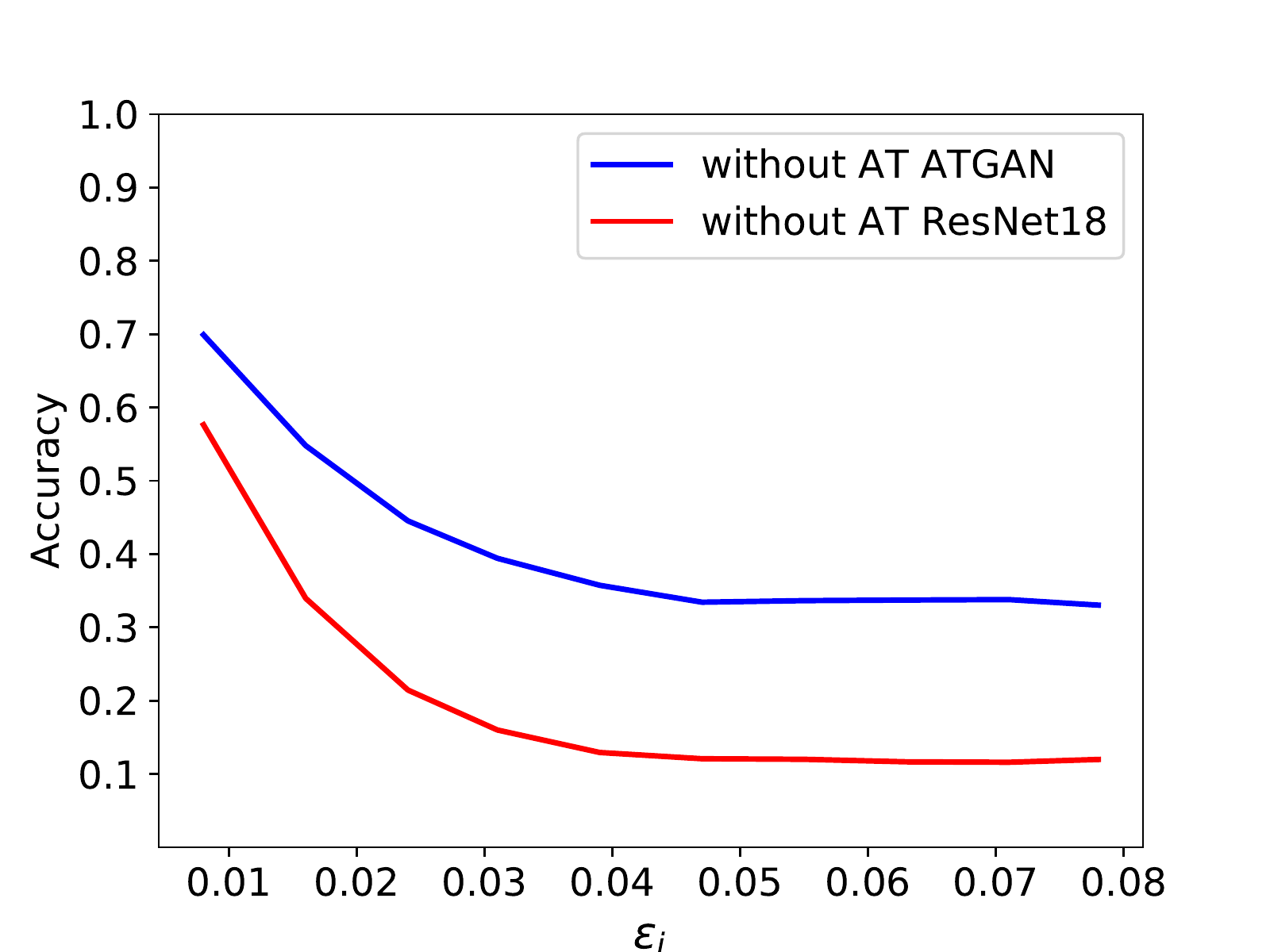}}
  \subfigure{\includegraphics[width=5.6cm]{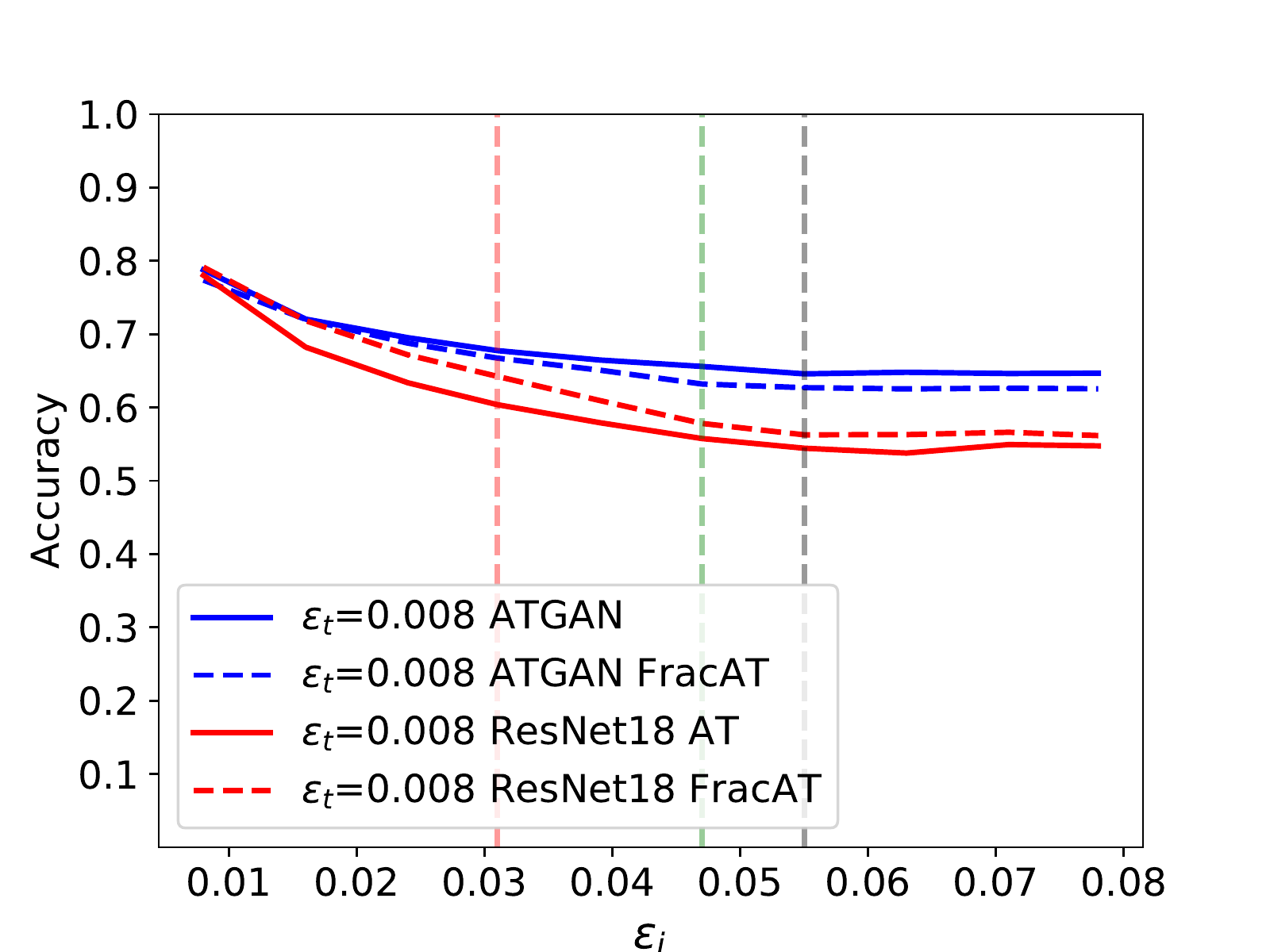}}
  \subfigure{\includegraphics[width=5.6cm]{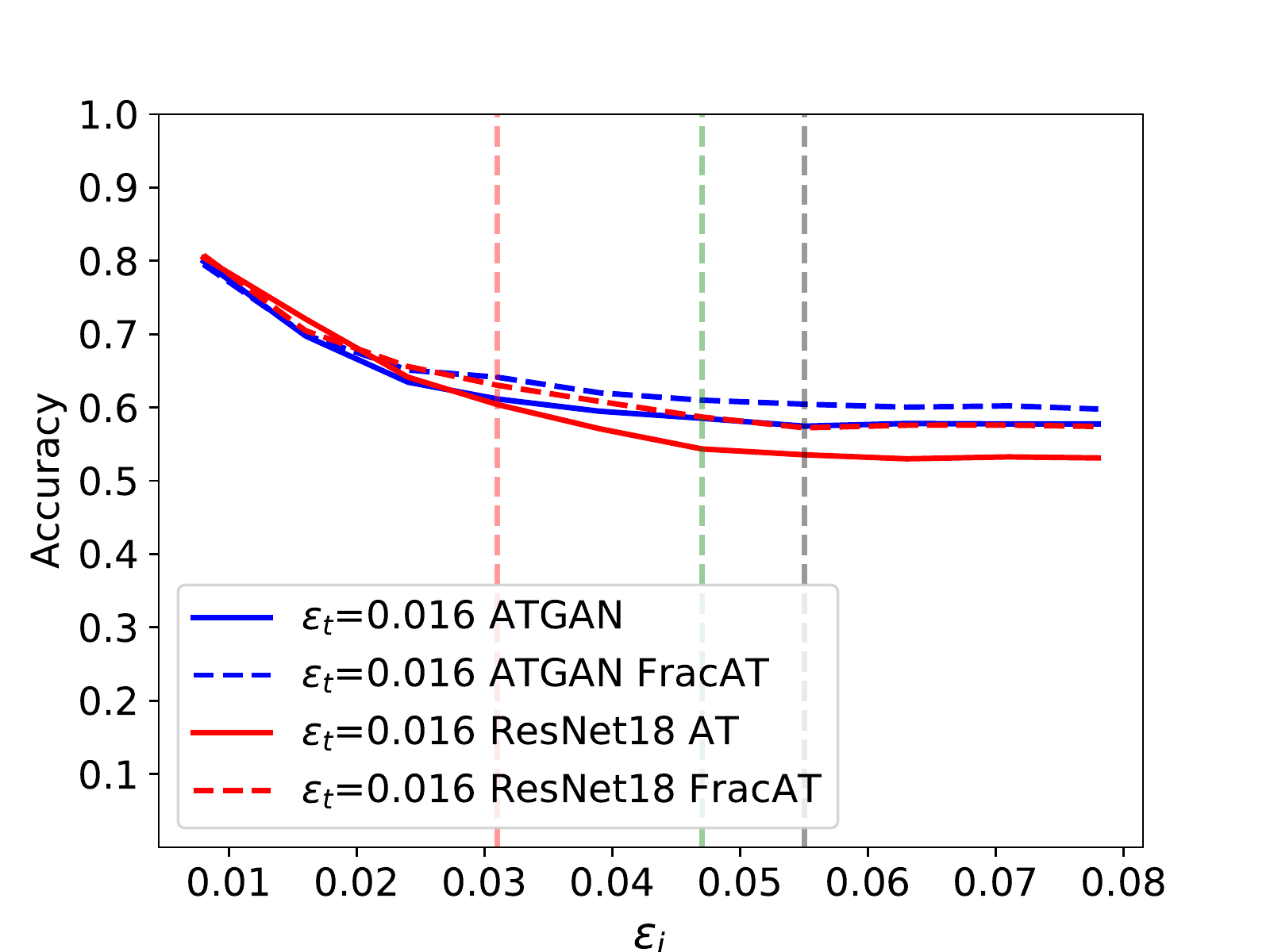}}
  \subfigure{\includegraphics[width=5.6cm]{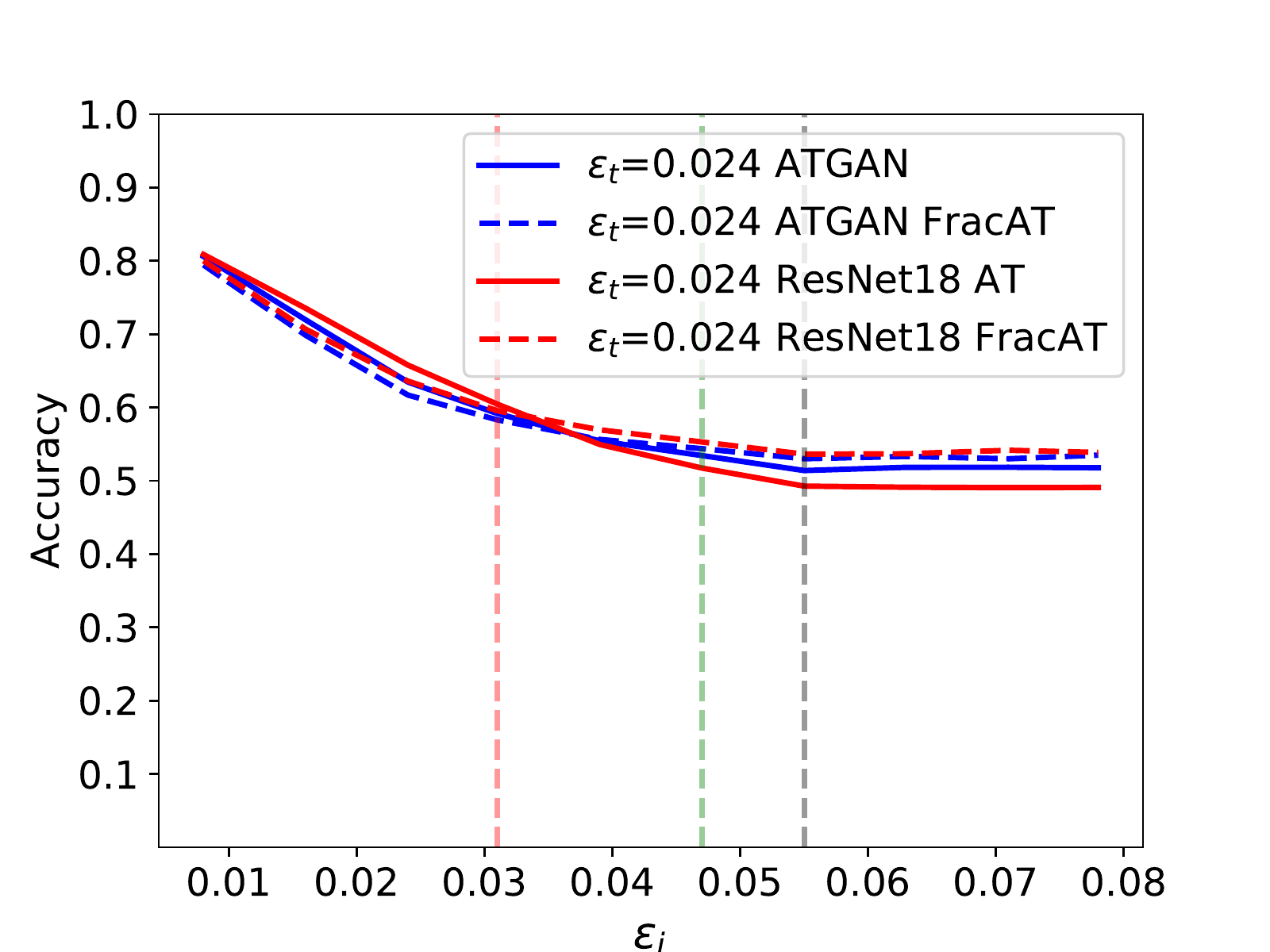}}
  \subfigure{\includegraphics[width=5.6cm]{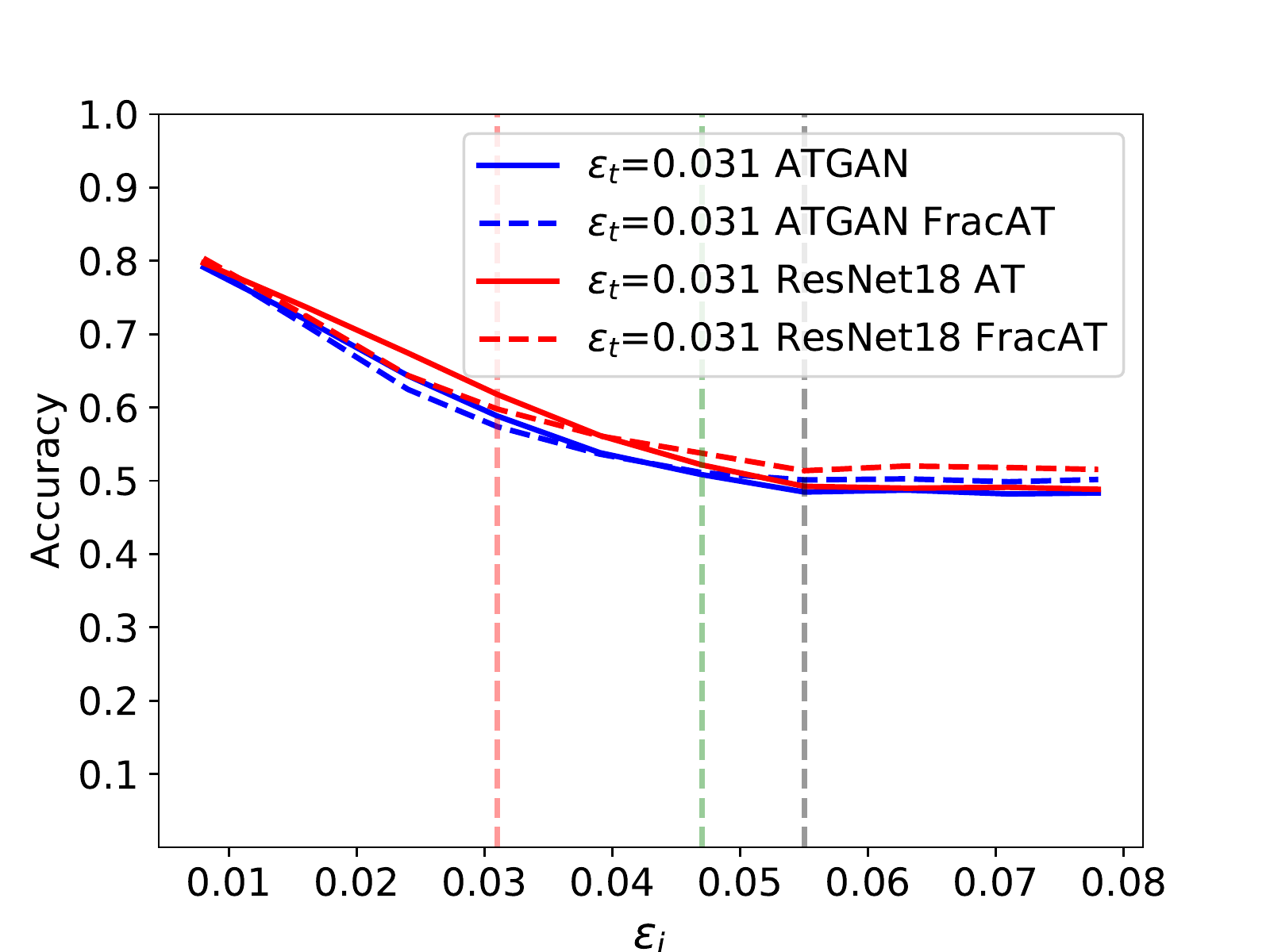}}
  \subfigure{\includegraphics[width=5.6cm]{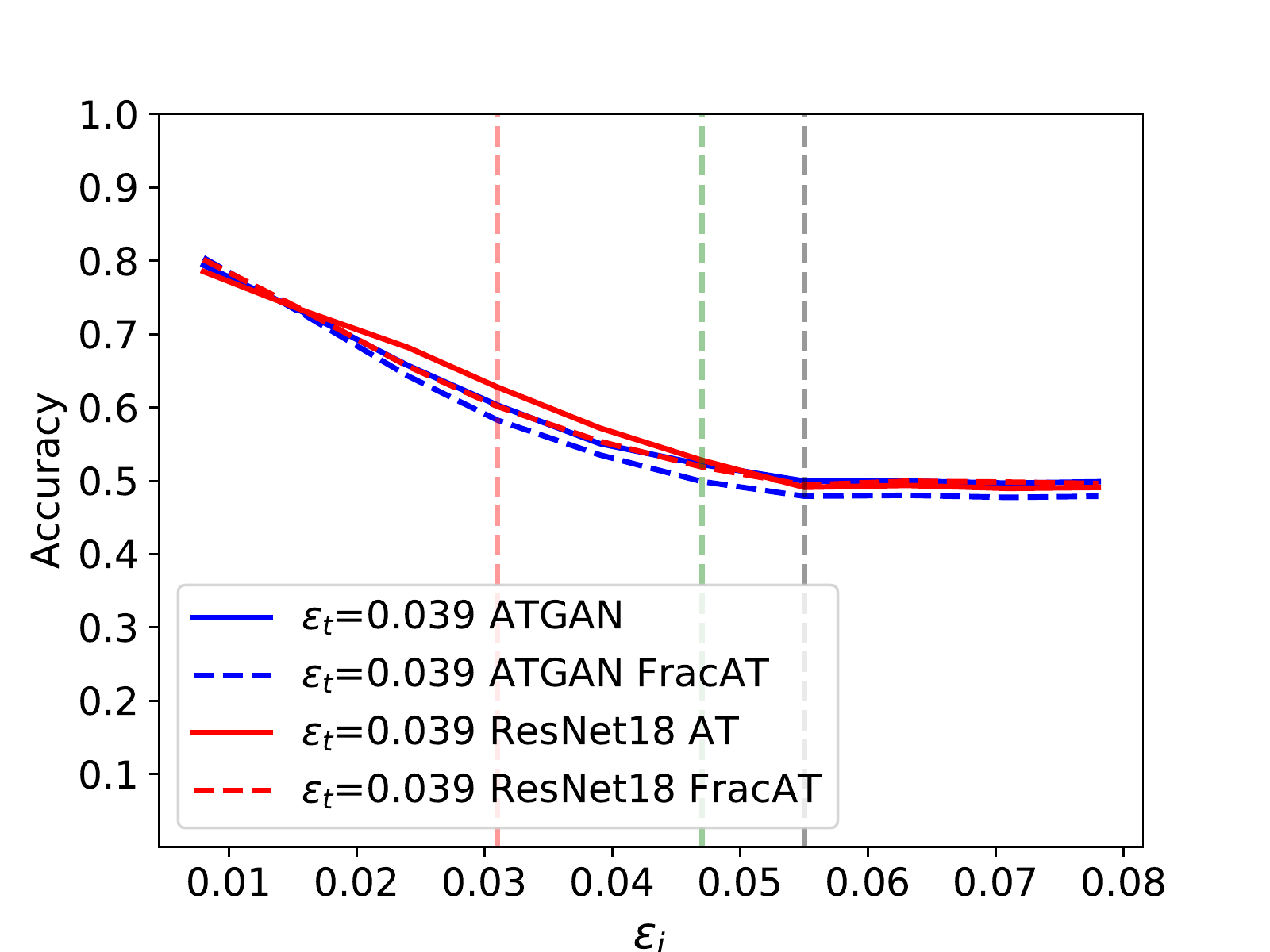}}
  \caption{\small The adversarial robustness generalization performance of ATGAN and ResNet18 on CIFAR-10 under white-box attack. The solid lines denote ATGAN. The dot-dash lines denote ResNet18}
  \label{Fig4}
\end{figure*}

\begin{figure*}[!htb]
  \centering
  \subfigure{\includegraphics[width=8cm]{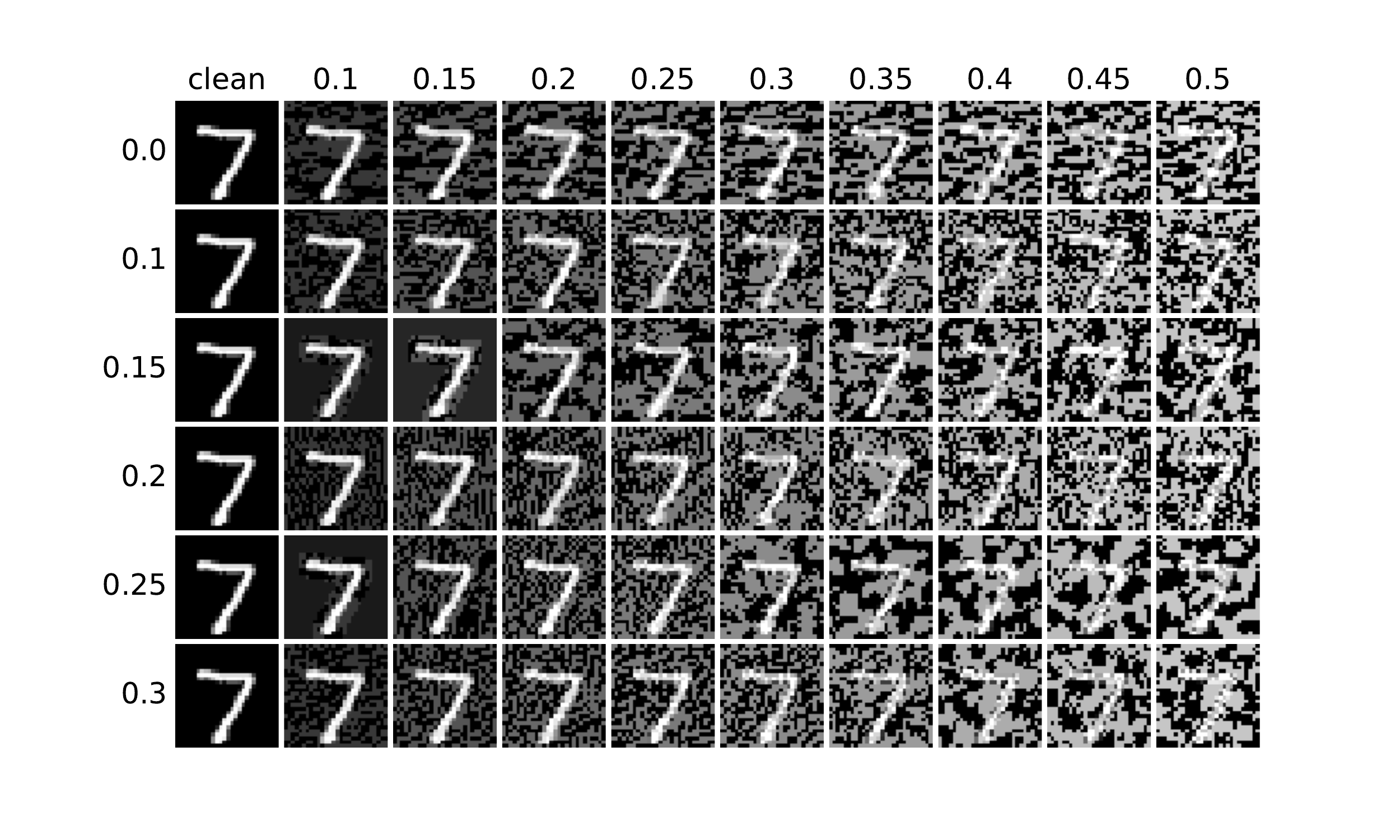}}
  \subfigure{\includegraphics[width=8cm]{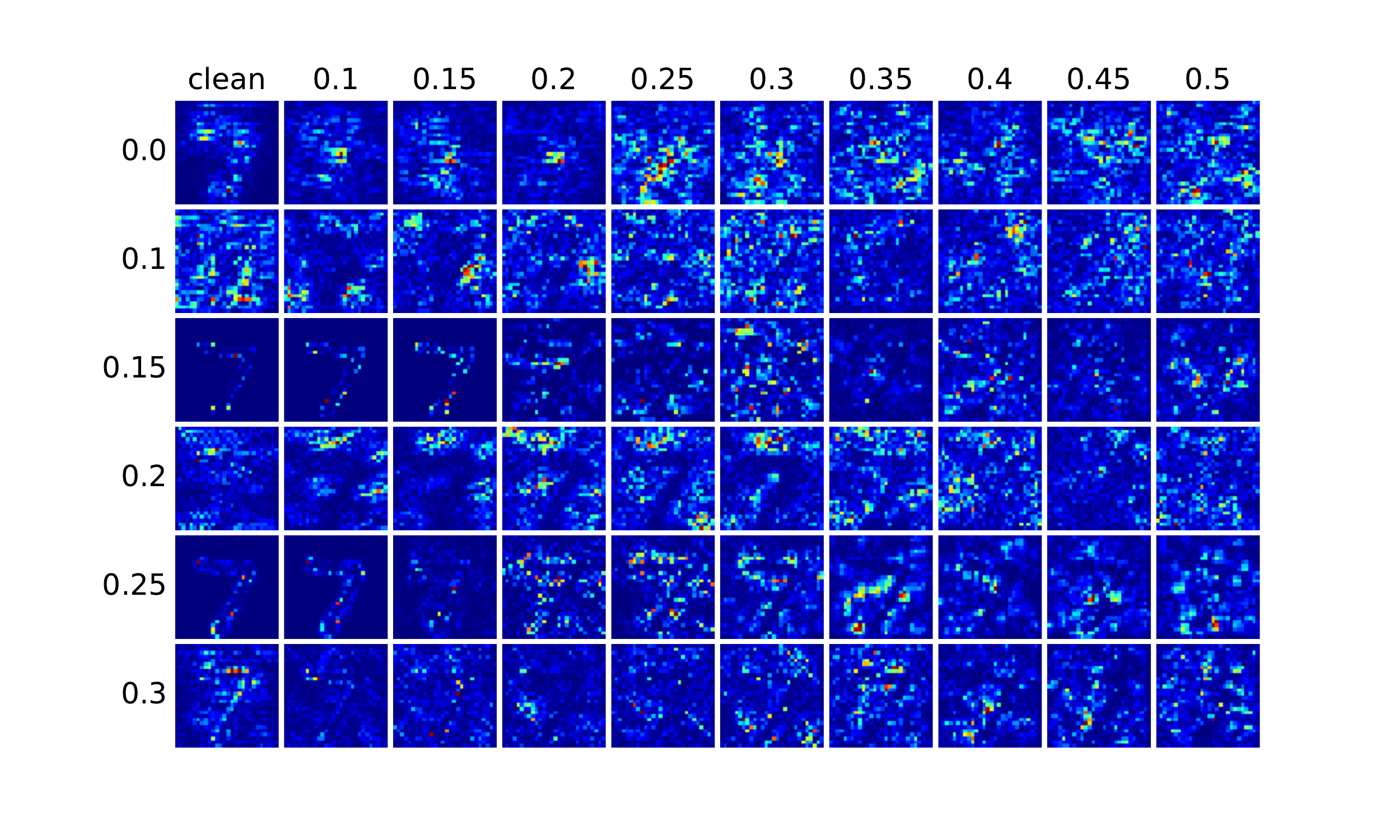}}
  \subfigure{\includegraphics[width=8cm]{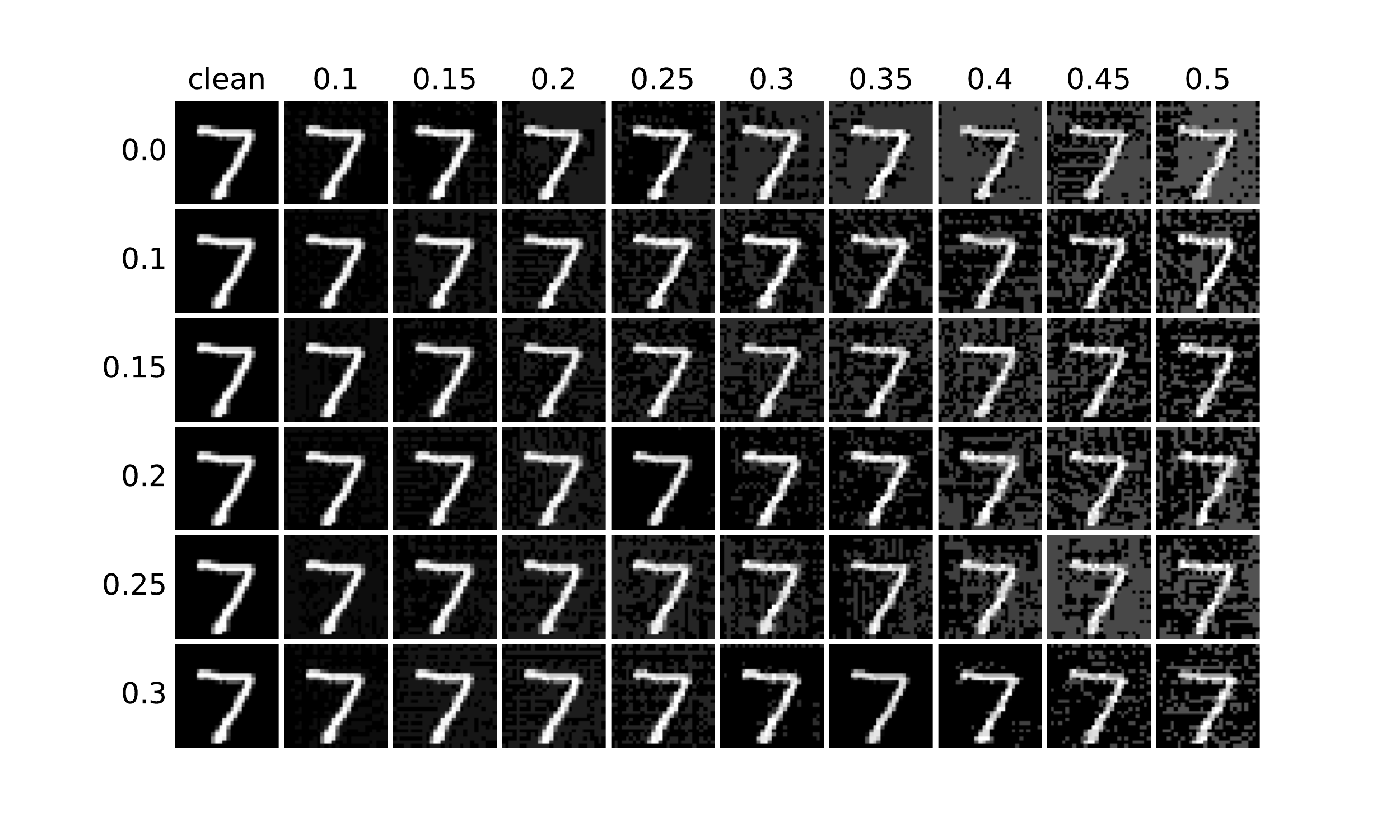}}
  \subfigure{\includegraphics[width=8cm]{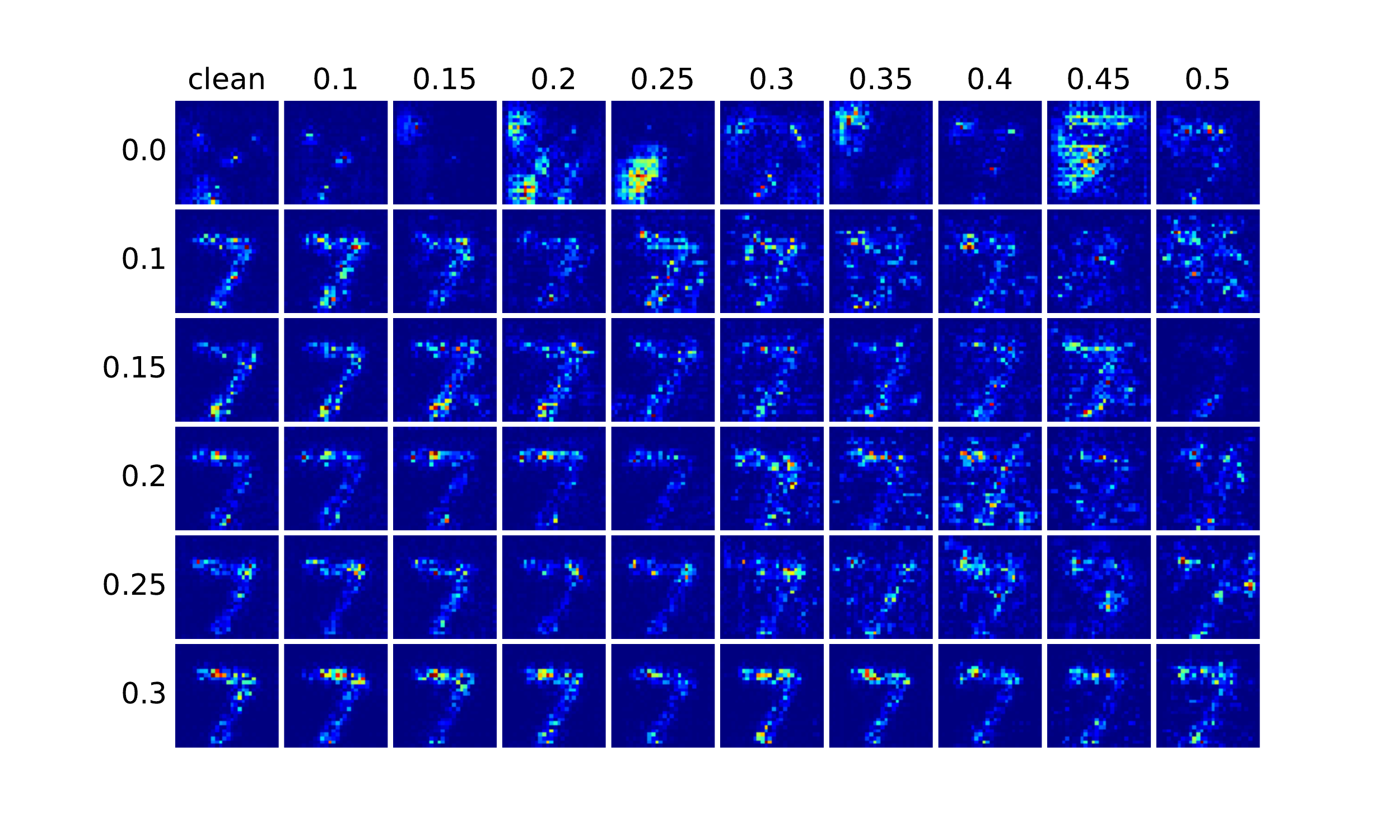}}
  \caption{\small Images and saliency maps of handwritten number '7' in MNIST dataset. Top left: the clean examples and the adversarial examples generated on the plain CNN; Top right: the saliency maps of the clean examples and the adversarial examples generated on the plain CNN; Bottom left: the clean examples and the adversarial examples generated on ATGAN; Bottom right: the saliency maps of the clean examples and the adversarial examples generated on ATGAN. The labels on the y-axis indicate $\epsilon_t$, where '0.0' indicates standard training. The labels on the x-axis indicate $\epsilon_i$, where 'clean' indicated the clean examples}
  \label{Fig5}
\end{figure*}

\begin{figure*}[!htb]
  \centering
  \subfigure{\includegraphics[width=8cm]{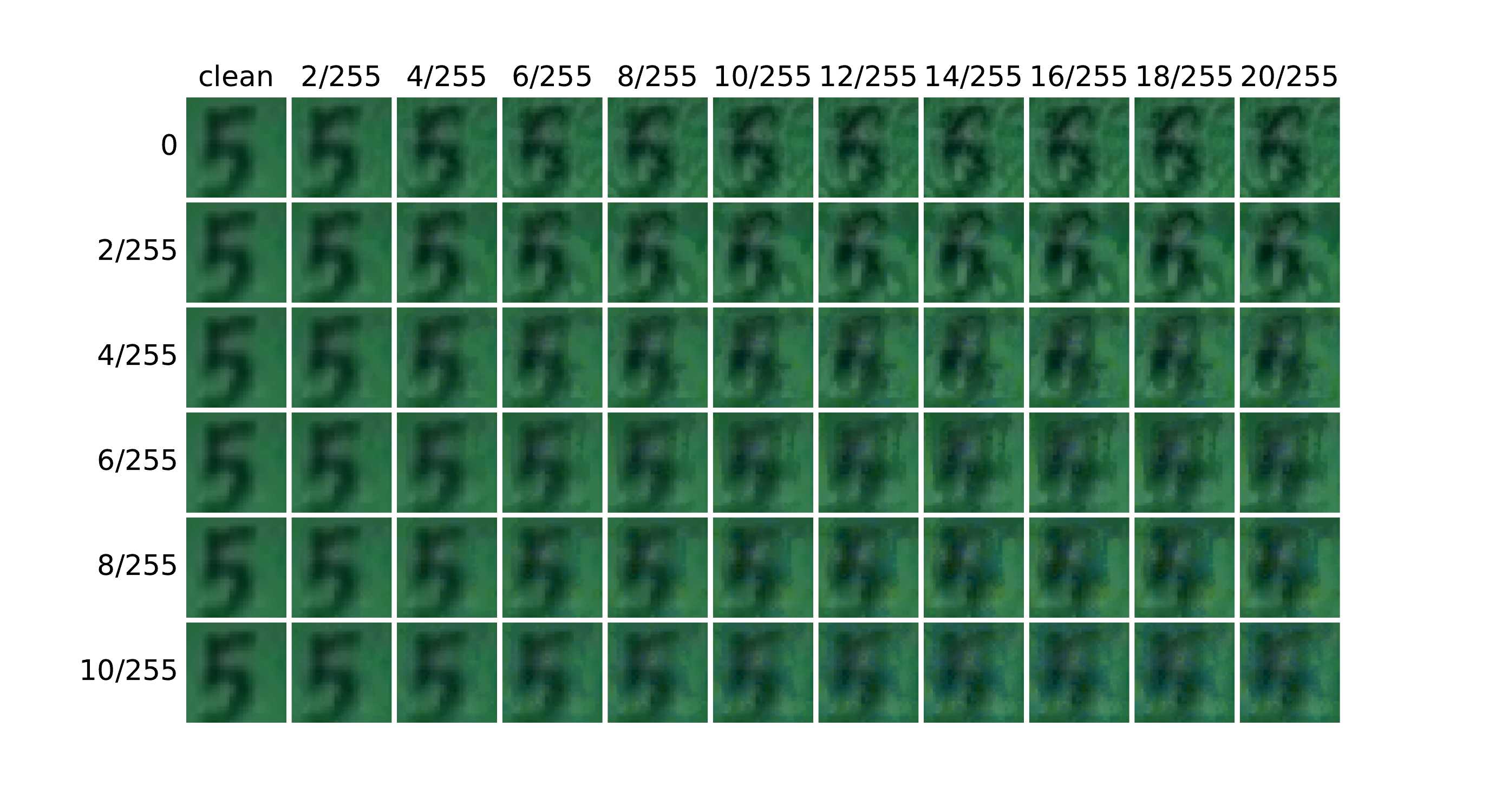}}
  \subfigure{\includegraphics[width=8cm]{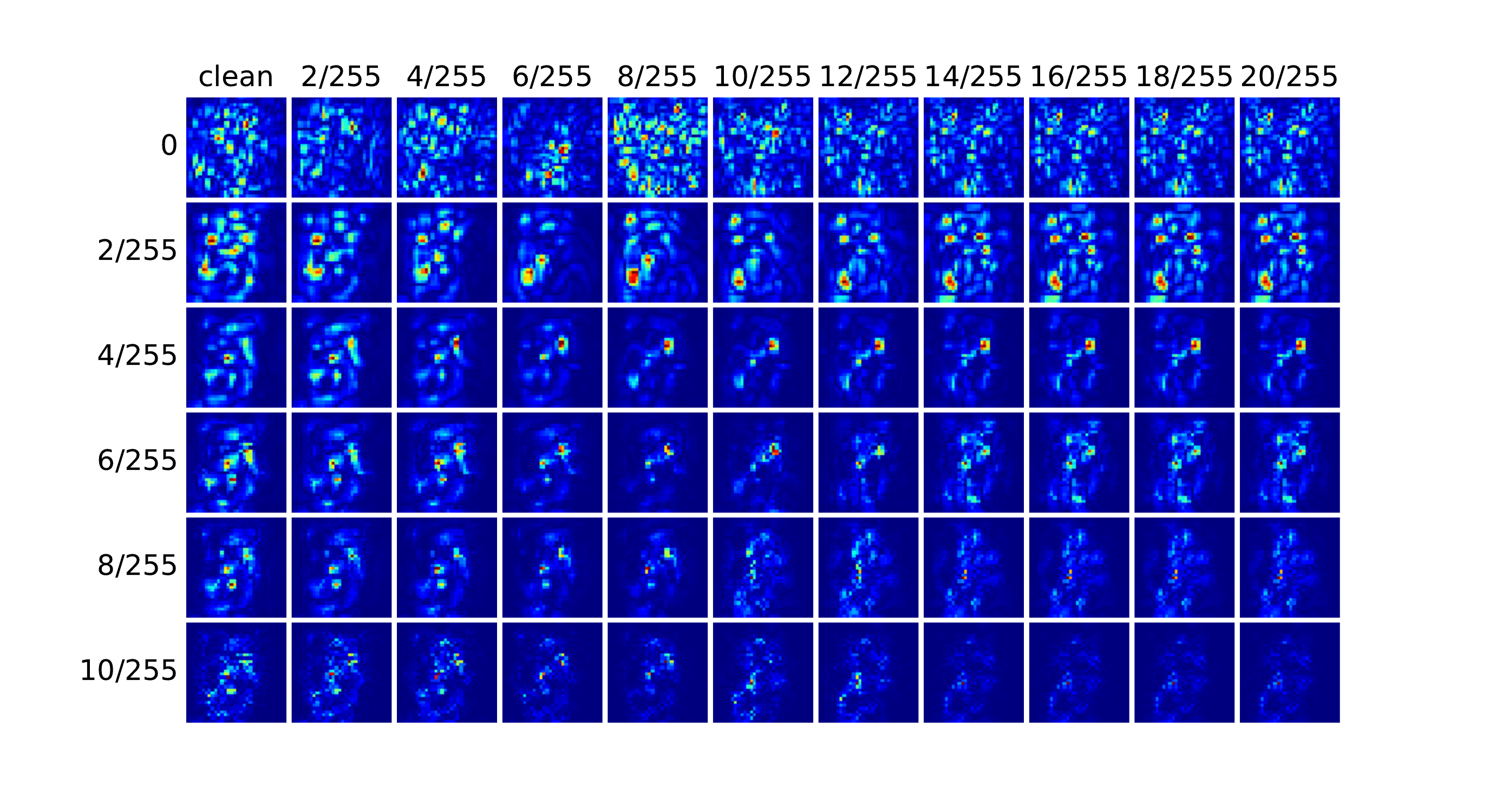}}
  \subfigure{\includegraphics[width=8cm]{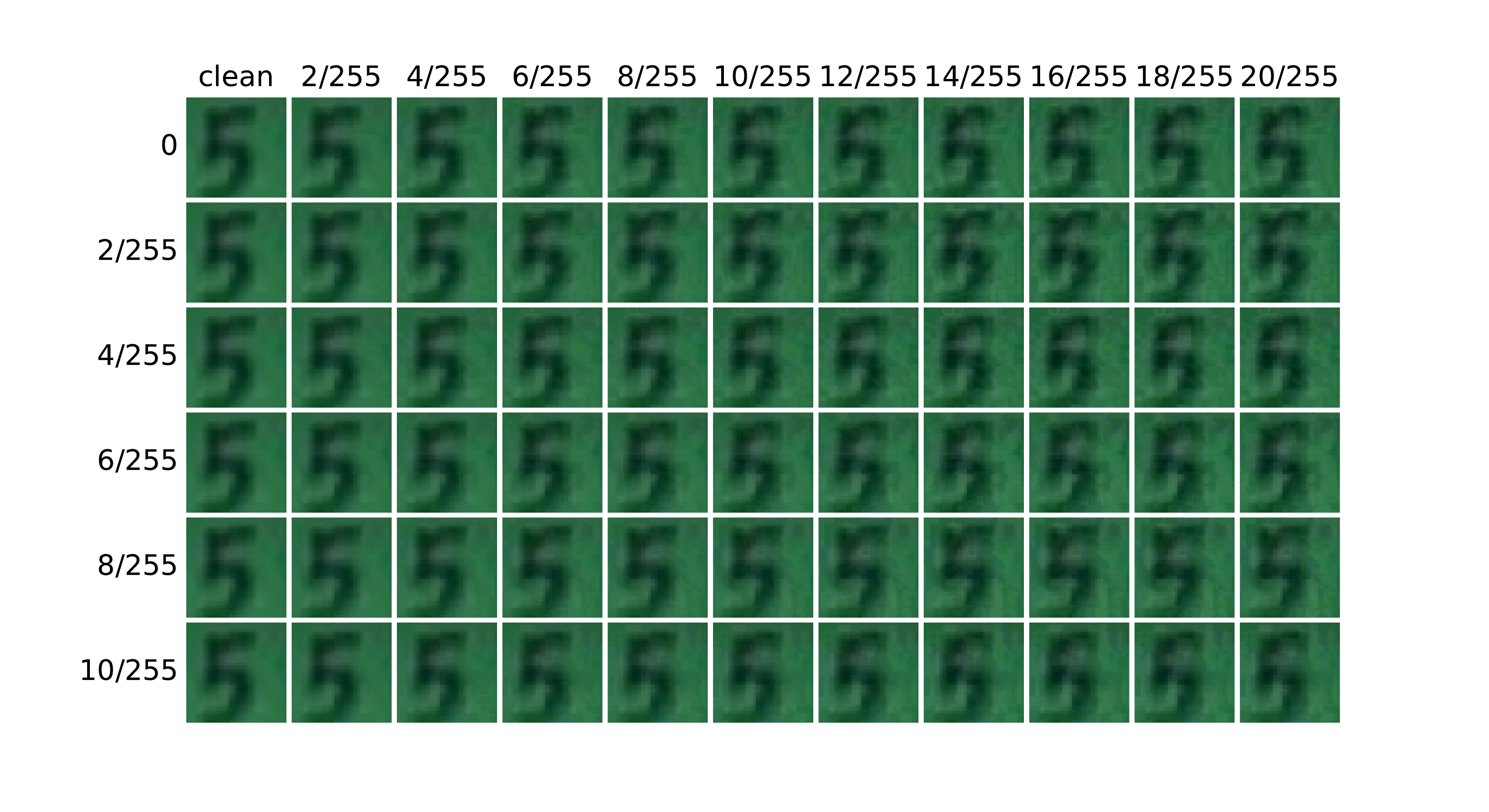}}
  \subfigure{\includegraphics[width=8cm]{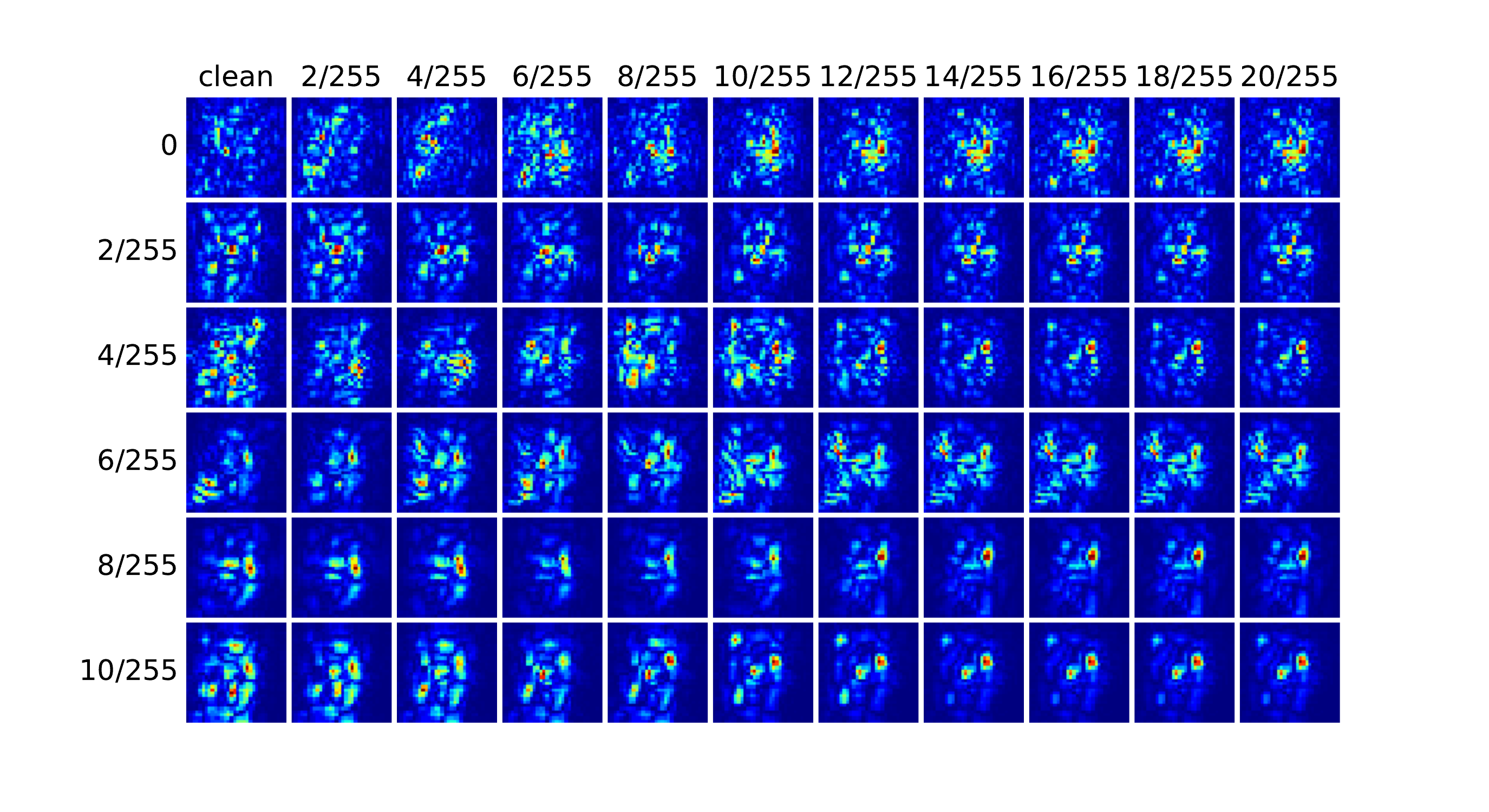}}
  \caption{\small Images and saliency maps of street view house number '5' in SVHN dataset. Top left: the clean examples and the adversarial examples generated on DenseNet40; Top right: the saliency maps of the clean examples and the adversarial examples generated on DenseNet40; Bottom left: the clean examples and the adversarial examples generated on ATGAN; Bottom right: the saliency maps of the clean examples and the adversarial examples generated on ATGAN. The labels on the y-axis indicate $\epsilon_t$, where '0.0' indicates standard training. The labels on the x-axis indicate $\epsilon_i$, where 'clean' indicated the clean examples}
  \label{Fig6}
\end{figure*}

\begin{figure*}[!htb]
  \centering
  \subfigure{\includegraphics[width=8cm]{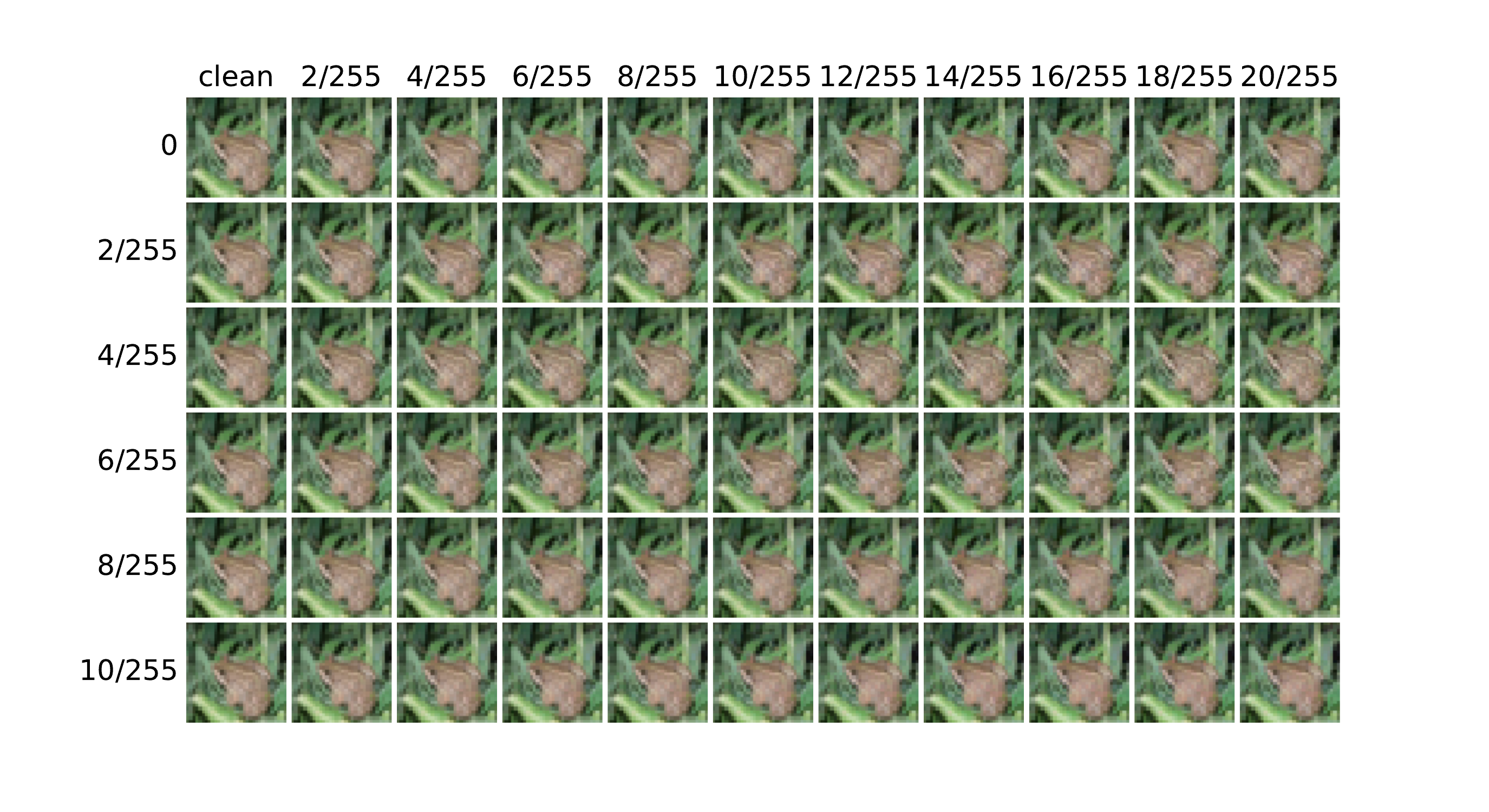}}
  \subfigure{\includegraphics[width=8cm]{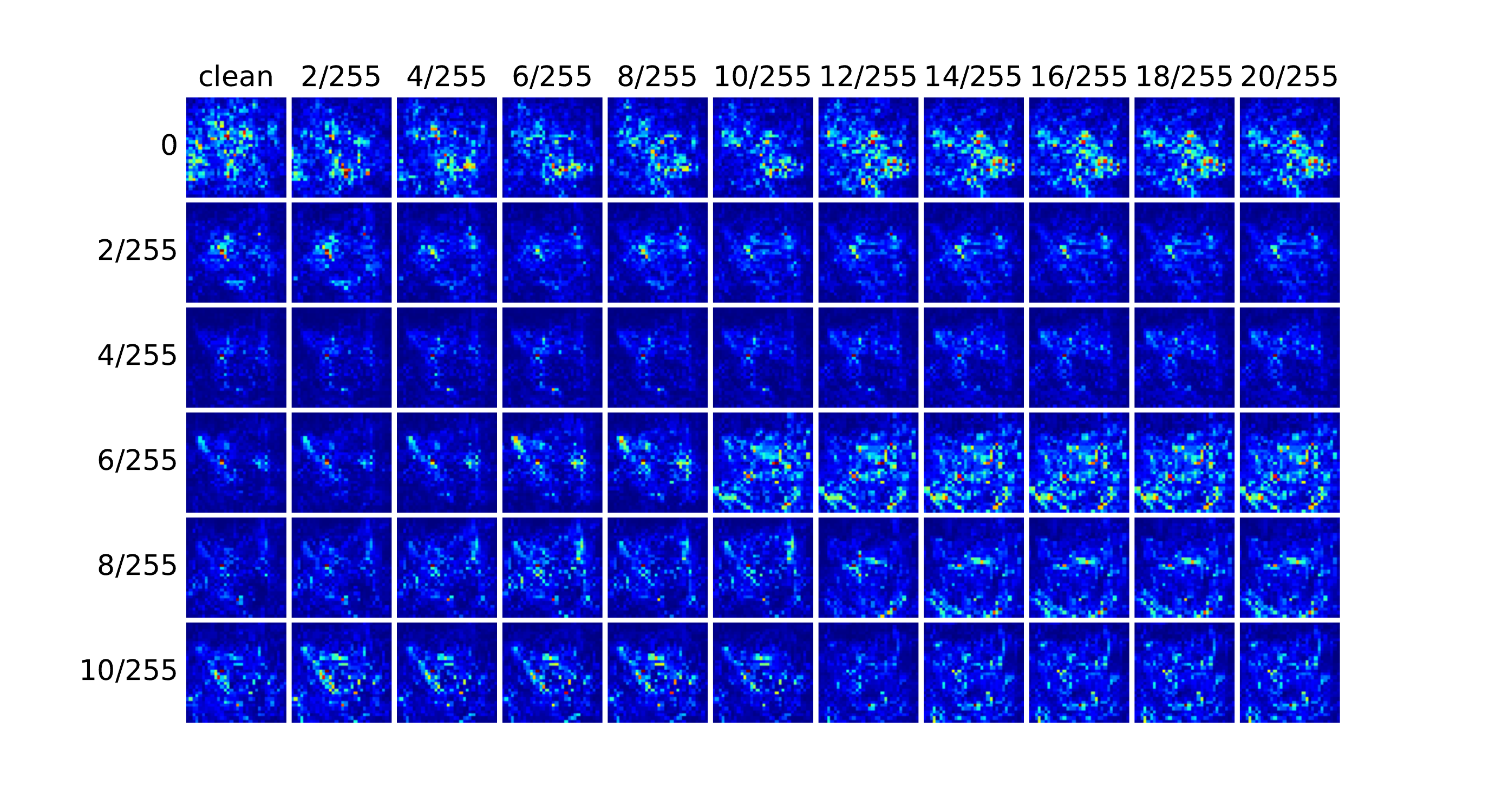}}
  \subfigure{\includegraphics[width=8cm]{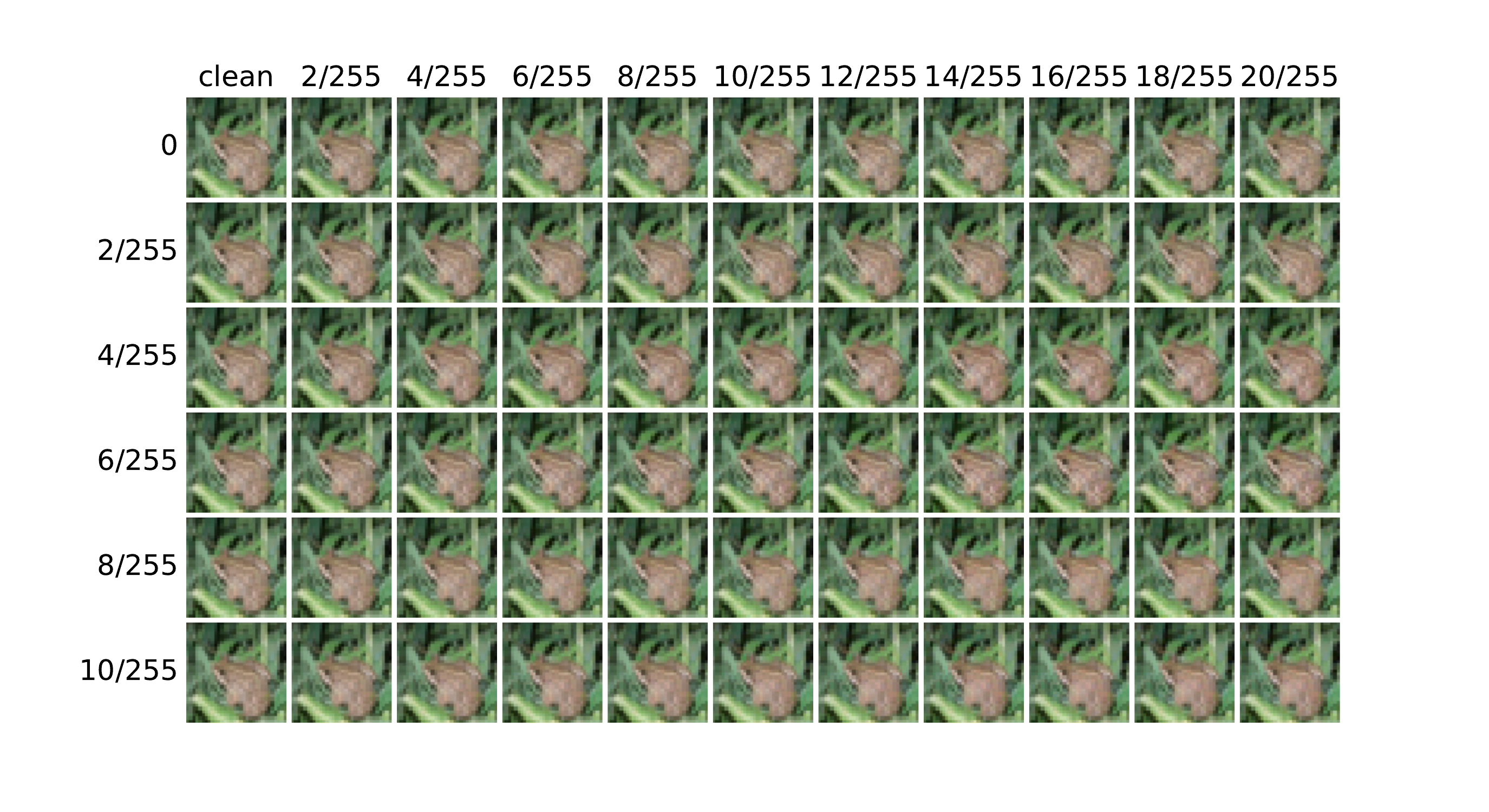}}
  \subfigure{\includegraphics[width=8cm]{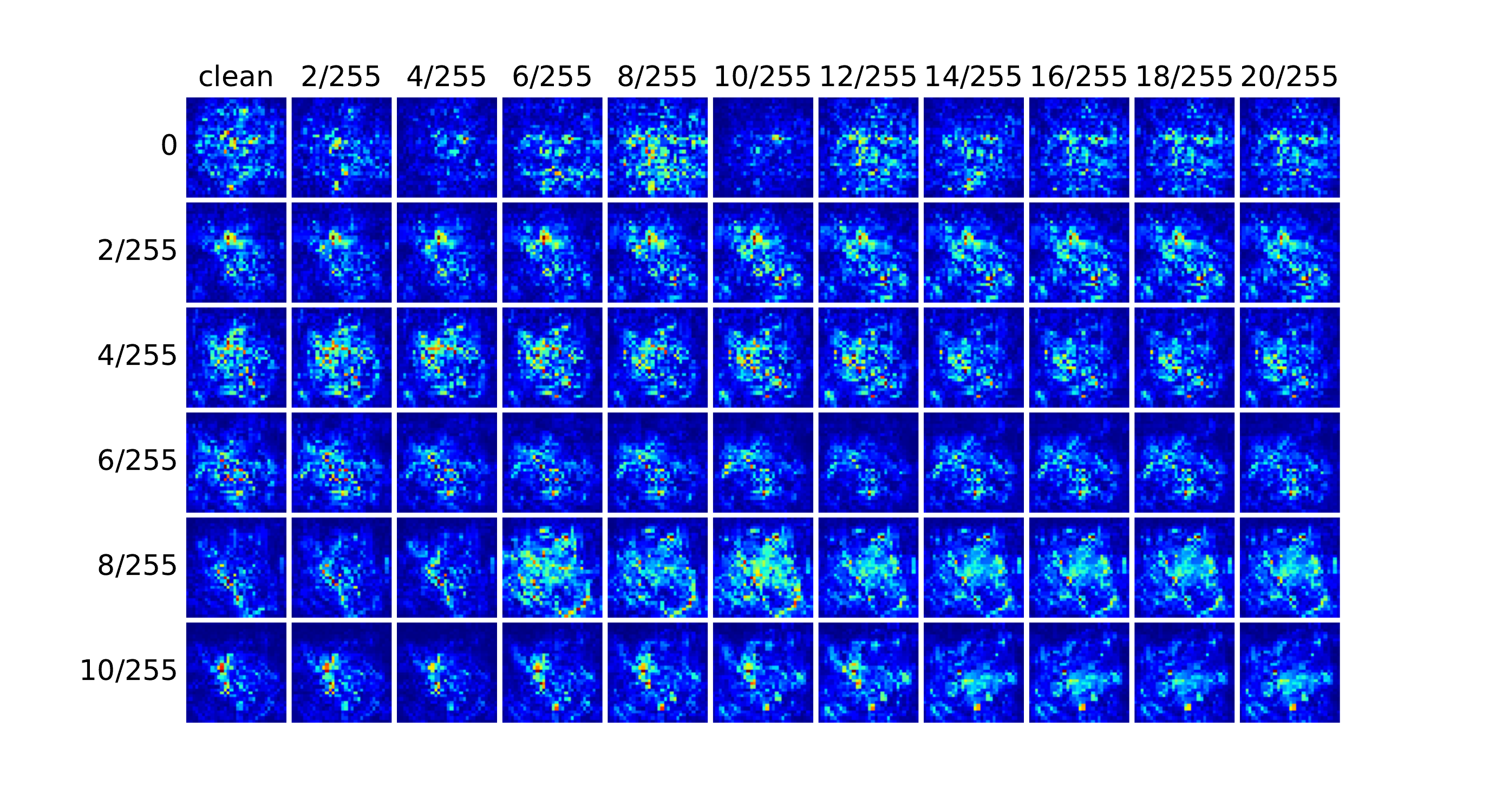}}
  \caption{\small Images and saliency maps of 'frog' in CIFAR-10 dataset. Top left: the clean examples and the adversarial examples generated on ResNet18; Top right: the saliency maps of the clean examples and the adversarial examples generated on ResNet18; Bottom left: the clean examples and the adversarial examples generated on ATGAN; Bottom right: the saliency maps of the clean examples and the adversarial examples generated on ATGAN. The labels on the y-axis indicate $\epsilon_t$, where '0.0' indicates standard training. The labels on the x-axis indicate $\epsilon_i$, where 'clean' indicated the clean examples}
  \label{Fig7}
\end{figure*}

\subsection{Results}
\label{sec:4.6}
\subsubsection{MNIST Experiments}
\label{sec:4.6.1}
We compare the adversarial robustness generalization performance of ATGAN with the plain CNN in the white-box attacks setting on the MNIST dataset using the threat model described above. The experimental results are shown in Fig. \ref{Fig2}. We can get three insights by analyzing Fig. \ref{Fig2}. First, the ATGAN without AT shows better adversarial robustness generalization performance than the plain CNN without AT. This indicates the superiority of the network architecture of ATGAN for resisting the adversarial examples on the MNIST dataset compared with the plain CNN. Second, after using AT, the adversarial robustness generalization performance of ATGAN and the plain CNN is further improved. The adversarial robustness accuracy of the plain CNN is higher than the ATGAN only when the magnitude of perturbation used for testing matches that used for training. When the magnitude of perturbation used for testing doesn't match that used for training, the adversarial robustness accuracy of the plain CNN is lower than the ATGAN and drops dramatically. By contrast, ATGAN show more robust adversarial generalization accuracy under different threat models. Finally, when using FracAT, the adversarial robustness generalization performance of ATGAN is still better than the plain CNN. 

In order to qualitatively evaluate and compare the adversarial robustness generalization performance of ATGAN and the plain CNN on the MNIST dataset, we compute the saliency maps of the clean examples and the adversarial examples, which highlight the pixels that matter most on the decision-making of the classification models. The results are exhibited in Fig. \ref{Fig5}. First, it is shown by the top left and the bottom left figures that the magnitude of the perturbation in the adversarial examples generated against ATGAN is much lower than that in the adversarial examples generated against the plain CNN. Second, for both ATGAN and the plain CNN that are trained by the standard procedure, the saliency maps don't exhibit discriminative features. After AT, the saliency maps exhibit more discriminative features, shown by the top right and bottom right features. Third, the pixels that matter most on the decision-making of the plain CNN trained by AT distribute in the background. As the strength of attacks increases, the discriminative features in the saliency maps become less discriminative. But for ATGAN, the most important pixels distribute in the foreground. As the strength of attacks increases, the discriminative features in the saliency maps stay discriminative. This indicated that ATGAN can learn image features that are more discriminative and robust than the plain CNN does on the MNIST dataset.

\subsubsection{SVHN Experiments}
\label{sec:4.6.2}
We compare the adversarial robustness generalization performance of ATGAN and DenseNet40 on the SVHN dataset under white-box attack. The experimental results are exhibited in Fig. \ref{Fig3}. We can obtain four insights by analyzing Fig. \ref{Fig3}. First, the adversarial robustness generalization performance of ATGAN is better than DenseNet40 when trained with standard procedure. This indicates the superiority of the network architecture of ATGAN for resisting the adversarial examples on the SVHN dataset compared with DenseNet40. Second, when using AT, the adversarial robustness generalization performance of both ATGAN and DenseNet40 are improved and the adversarial robustness generalization performance of ATGAN is better than DenseNet40 for all $\epsilon_t$. Third, for both ATGAN and DenseNet40, the adversarial robustness generalization performance increases as $\epsilon_t$ increases. Finally, models trained with FracAT show different global adversarial robustness generalization performance rank for different $\epsilon_t$, when $\epsilon_t=2/255$, $4/255$, and $6/255$, the adversarial robustness generalization accuracies of ATGAN are higher than DenseNet40, when $\epsilon_t=8/255$ and $10/255$, the robustness curves of ATGAN and DenseNet40 intersect.

In order to qualitatively evaluate and compare the adversarial robustness generalization performance of ATGAN and DenseNet40 on the SVHN dataset, we compute the saliency maps of the clean examples and the adversarial examples. The results are exhibited in Fig. \ref{Fig6}. First, it is shown by the top left and the bottom left figures that the semantic information in the adversarial examples generated against DenseNet40 changes from '5' to '6' as the strength of attack increases. But the semantic information in the adversarial examples generated against ATGAN stays unchanged. Second, the pixels that matter most on the decision-making of ATGAN trained by standard procedure exhibit more concentrated patterns in the saliency maps compared to DenseNet40, shown by the first rows in the top right and bottom right figures. Third, the foreground is more important to the decision-making for ATGAN compared to DenseNet40, which is indicated by the higher contrast between the foreground and background of the saliency maps.

\begin{table*}[!htb]
  \centering
  \caption{\small Point-wise measures for the plain CNN and ATGAN trained by AT on MNIST. Each row contains the adversarial robustness generalization accuracies of different models trained by AT with different perturbation thresholds then evaluated under one point-wise measure. The darker tune of gray in the cell indicates better adversarial robustness generalization performance. 'A' 'AFracAT' 'pCNN' and 'PFracAT' stand for 'ATGAN' 'ATGAN FracAT' 'plain CNN' and 'plain CNN FracAT'}
  \label{table2}
  \begin{tabular}{c|cccc|cccc}
    \hline
    \multirow{2}{*}    &     \multicolumn{4}{c}{$\epsilon_t=$ 0.2}   &     \multicolumn{4}{|c}{$\epsilon_t=$ 0.3}    \\ \cline{2-9}
                       &   A   &   AFracAT   &   pCNN   &   PFracAT  &   A   &   AFracAT   &   pCNN   &   PFracAT    \\ \hline
    $\epsilon_i=$ 0.15 & \cellcolor{gray!50}0.981 & 0.980 & \cellcolor{gray!100}0.999 & \cellcolor{gray!75}0.990 & \cellcolor{gray!100}0.983 & \cellcolor{gray!75}0.978 & 0.857 & \cellcolor{gray!50}0.971 \\ 
    $\epsilon_i=$ 0.30 & \cellcolor{gray!75}0.959 & \cellcolor{gray!100}0.966 & \cellcolor{gray!50}0.827 & 0.807 & \cellcolor{gray!50}0.970 & \cellcolor{gray!50}0.970 & \cellcolor{gray!100}0.999 & \cellcolor{gray!75}0.989 \\ 
    $\epsilon_i=$ 0.45 & \cellcolor{gray!100}0.952 & \cellcolor{gray!75}0.948 & 0.151 & \cellcolor{gray!50}0.193 & \cellcolor{gray!75}0.952 & \cellcolor{gray!100}0.960 & 0.050 & \cellcolor{gray!50}0.059 \\ \hline
  \end{tabular}
\end{table*}

\begin{table*}[!htb]
  \centering
  \caption{\small Point-wise measures for DenseNet40 and ATGAN trained by AT on SVHN. Each row contains the adversarial robustness generalization accuracies of different models trained by AT with different perturbation thresholds then evaluated under one point-wise measure. The darker tune of gray in the cell indicates better adversarial robustness generalization performance. 'A' 'AFracAT' 'D' and 'DFracAT' stand for 'ATGAN' 'ATGAN FracAT' 'DenseNet40' and 'DenseNet40 FracAT'}
  \label{table3}
  \begin{tabular}{c|cccc|cccc}
    \hline
    \multirow{2}{*}       &     \multicolumn{4}{c}{$\epsilon_t=$ 6/255} &     \multicolumn{4}{|c}{$\epsilon_t=$ 8/255}  \\ \cline{2-9}
                          &    A   &   AFracAT   &   D   &    DFracAT   &    A   &    AFracAT   &    D    &    DFracAT  \\ \hline
     $\epsilon_i=$ 8/255  & \cellcolor{gray!100}0.634 & \cellcolor{gray!50}0.561 & \cellcolor{gray!75}0.570 & 0.536 & \cellcolor{gray!100}0.680 & \cellcolor{gray!75}0.620 & 0.579 & \cellcolor{gray!50}0.590 \\ 
     $\epsilon_i=$ 12/255 & \cellcolor{gray!100}0.454 & \cellcolor{gray!50}0.380 & \cellcolor{gray!75}0.386 & 0.341 & \cellcolor{gray!100}0.518 & \cellcolor{gray!50}0.421 & 0.403 & \cellcolor{gray!75}0.452 \\ 
     $\epsilon_i=$ 14/255 & \cellcolor{gray!100}0.387 & \cellcolor{gray!75}0.323 & \cellcolor{gray!50}0.319 & 0.272 & \cellcolor{gray!100}0.454 & \cellcolor{gray!50}0.345 & 0.338 & \cellcolor{gray!75}0.410 \\ \hline
  \end{tabular}
\end{table*} 

\begin{table*}[!htb]
  \centering
  \caption{\small Point-wise measures for ResNet18 and ATGAN trained by AT on CIFAR-10. Each row contains the adversarial robustness generalization accuracies of different models trained by AT with different perturbation thresholds then evaluated under one point-wise measure. The darker tune of gray in the cell indicates better adversarial robustness generalization performance. 'A' 'AFracAT' 'R' and 'RFracAT' stand for 'ATGAN' 'ATGAN FracAT' 'ResNet18' and 'ResNet18 FracAT'}
  \label{table4}
  \begin{tabular}{c|cccc|cccc}
    \hline
    \multirow{2}{*}      &     \multicolumn{4}{c}{$\epsilon_t=$ 2/255} &     \multicolumn{4}{|c}{$\epsilon_t=$ 6/255}  \\ \cline{2-9}
                         &    A   &    AFracAT   &   R   &   RFracAT   &    A   &    AFracAT   &   R   &    RFracAT    \\ \hline
    $\epsilon_i=$ 8/255  & \cellcolor{gray!100}0.678 & \cellcolor{gray!75}0.667 & 0.604 & \cellcolor{gray!50}0.642 & \cellcolor{gray!50}0.592 & 0.583 & \cellcolor{gray!100}0.604 & \cellcolor{gray!75}0.595 \\ 
    $\epsilon_i=$ 12/255 & \cellcolor{gray!100}0.656 & \cellcolor{gray!75}0.632 & 0.558 & \cellcolor{gray!50}0.578 & \cellcolor{gray!50}0.534 & \cellcolor{gray!75}0.544 & 0.517 & \cellcolor{gray!100}0.553 \\ 
    $\epsilon_i=$ 14/255 & \cellcolor{gray!100}0.646 & \cellcolor{gray!75}0.627 & 0.544 & \cellcolor{gray!50}0.563 & \cellcolor{gray!50}0.514 & \cellcolor{gray!75}0.530 & 0.493 & \cellcolor{gray!100}0.536 \\ \hline
  \end{tabular}
\end{table*}

\subsubsection{CIFAR-10 Experiments}
\label{sec:4.6.3}
We also compare the adversarial robustness generalization performance of ATGAN and ResNet18 on the CIFAR-10 dataset under white-box attacks. The experimental results are shown in Fig. \ref{Fig4}. We can get four insights by analyzing Fig. \ref{Fig4}. First, when trained using standard procedure, the adversarial robustness generalization performance of ATGAN is better than ResNet18 by a large margin. This indicates the superiority of the network architecture of ATGAN for resisting the adversarial examples on CIFAR-10 compared with ResNet18. Second, when trained using AT, the adversarial robustness generalization performance of both ATGAN and ResNet18 improves. For  $\epsilon_t=2/255$, the adversarial robustness generalization accuracies of ATGAN are higher than ResNet18 for all $\epsilon_i$. For $\epsilon_t=8/255$, the adversarial robustness generalization accuracies of ATGAN are lower than ResNet18 for all $\epsilon_i$. For other $\epsilon_t$, the robustness curves of ATGAN and ResNet18 intersect. This indicates a trade-off between ATGAN and ResNet18 in terms of adversarial robustness generalization performance on CIFAR-10. Third, for both ATGAN and ResNet18, the adversarial robustness generalization performance decreases as $\epsilon_t$ increases. Finally, a similar trade-off is also observed on the models trained with FracAT.

In order to qualitatively evaluate and compare the adversarial robustness generalization performance of ATGAN and ResNet18 on the CIFAR-10 dataset, we compute the saliency maps of the clean examples and the adversarial examples. The results are exhibited in Fig. \ref{Fig7}. For the more complicated dataset, the perturbations in the adversarial examples are imperceptible to human eyes. But the saliency maps exhibit different features for different examples and different models. First, both ATGAN and ResNet18 learn more discriminative features after AT. Second, similar to MNIST and SVHN datasets, the foregrounds matter more on the decision-making of ATGAN compared to ResNet18.

\subsubsection{Obfuscated Gradients Test}
\label{sec:4.6.4}
The evaluation of adversarial defense methods should be careful because obfuscated gradients are a common occurrence which leads to a false sense of security in defending against adversarial examples. Following the advise of Athalye \textit{et al.} \cite{40}, we conduct two experiments to test whether ATGAN rely on obfuscated gradients or not. We attack ATGAN with 1000-step PGD with $\epsilon_i=0.3$ for MNIST and $\epsilon_i=8/255$ for SVHN and CIFAR-10. And we attack ATGAN with 100-step PGD with $\epsilon_i=\infty$. Without loss of generality, we choose the ATGAN trained with $\epsilon_t=0.3$ for MNIST and the ATGAN trained with $\epsilon_t=8/255$ for SVHN and CIFAR-10. The experimental results are exhibited in Table \ref{table5}. ATGAN achieves similar robust accuracy on PGD-1K compared to PGD-40 for MNSIT and PGD-7 for SVHN and CIFAR-10. And the robust accuracy becomes zero on unconstrained PGD-100 attack. This indicates that the adversarial robustness of ATGAN is not a result of obfuscated gradients.

\begin{table}[h]
  \centering
  \caption{\small Obfuscated Gradients Test. PGD denotes 40-step PGD with $\epsilon_i=0.3$ for MNIST and 7-step PGD with $\epsilon_i=8/255$ for SVHN and CIFAR-10. PGD-1k denotes 1000-step PGD}
  \label{table5}
  \begin{tabular}{cccc}
    \hline
    Dataset  & PGD   & PGD-1k & $\epsilon_i=\infty$ \\ \hline
    MNIST    & 0.970 & 0.970  & 0.005               \\
    SVHN     & 0.680 & 0.642  & 0.000               \\
    CIFAR-10 & 0.588 & 0.560  & 0.007               \\
    \hline
  \end{tabular}
\end{table}

\section{Discussions}
\label{sec:5}
\textbf{Local vs Global Perspective of Adversarial Robustness} On the comparison of the adversarial robustness of different models, the most common method is the point-wise measure: pick up some perturbation threshold then compare the robust accuracy of different models based on the perturbation threshold. And for AT, most recent publications only use a single perturbation threshold during training. These methods have the problem of reflecting global properties concerning the adversarial robustness of different models. Our experimental results reflect this notion. The vertical dashed lines plotted in Figure \ref{Fig2}, \ref{Fig3}, and \ref{Fig4} correspond to different $\epsilon_i$. We take the adversarial accuracies at the intersection points of the adversarial curves and the vertical dashed lines and compared them in Table \ref{table2}, \ref{table3}, and \ref{table4}. According to Table \ref{table2}, \ref{table3}, and \ref{table4}, the models trained by AT with different perturbation thresholds exhibit different adversarial robustness when evaluated under the same point-wise measure. And the models trained by AT with the same perturbation threshold exhibit different adversarial robustness under different point-wise measures. This observation indicates that the point-wise measure can only provide a local perspective of adversarial robustness. To comprehensively compare the adversarial robustness of different models and get a global perspective, the robustness curves should be leveraged as a tool.

\textbf{Superiority of ATGAN in Architecture} The experimental results exhibited in Section 4.8 show that the magnitude of perturbations in the adversarial examples generated against ATGAN is much lower than that in the adversarial examples generated against the baseline models, especially for MNIST and SVHN datasets. This is a qualitative performance of the superiority of adversarial robustness generalization of ATGAN in architecture. And this benefit comes from combining the AT with the GANs training procedure. During the training of the GANs, the discriminator loss and the perceptual loss encourage the distribution of the output of the generator to be similar to the clean examples. This objective has not only the benefit of denoising but also the benefit of encouraging the AT to generate adversarial examples with a lower magnitude of perturbations, which restrict the power of the adversarial examples. On the other hand, the discriminator loss and the perceptual loss don't drive the output of the generator to match the clean examples exactly, making the generator play the role of data augmentation that increases the sample complexity of the adversarial examples, which also improve the adversarial robustness generalization. 

\textbf{Specificity of Adversarial Robustness in terms of Datasets} By analyzing the robustness curves of different datasets, we find that the impact of $\epsilon_t$ and $\epsilon_i$ on the adversarial robustness generalization performance of ATGAN has unique patterns in terms of different datasets. For the MNIST dataset, ATGAN trained by AT with different $\epsilon_t$ exhibit comparable adversarial robustness generalization performance, therefore we don't need to pay too much attention to the choice of $\epsilon_t$ when adversarially training ATGAN on MNIST. For the SVHN dataset, the adversarial robustness generalization performance of ATGAN increases as $\epsilon_t$ increases, this indicates that we need to choose a high value of $\epsilon_t$ when adversarially training the ATGAN on SVHN. And for CIFAR-10, different from MNIST and SVHN, the adversarial robustness generalization performance of ATGAN decreases as $\epsilon_t$ increases, this indicates that we need to choose a low value of $\epsilon_t$ when adversarially training ATGAN. By the discussion above, we can get the notion that the distribution of the dataset has a significantly impact on the configuration of parameters when adversarially training ATGAN. So we need to leverage different configurations when applying ATGAN to different datasets.

\section{Conclusions}
\label{sec:6}
In order to defend against adversarial examples, this paper proposes ATGAN that incorporates AT into standard GANs training procedure to gain robustness against $L_{\infty}$ norm constrained perturbations. Experimental results demonstrate that ATGAN not only outperforms the state-of-the-art CNN classifiers trained by AT in terms of adversarial robustness generalization performance but also overcomes the obfuscated gradients problems of existing generative model-based adversarial defense methods. Moreover, ATGAN can learn more discriminative and robust features than the baseline models, which can explain the superiority of ATGAN in adversarial robustness generalization performance.

\end{document}